\documentclass{bmvc2k}

\usepackage{graphicx}
\usepackage{amsfonts}
\usepackage{pifont}
\usepackage[table]{xcolor}
\usepackage{amsmath}
\usepackage{tcolorbox}
\usepackage{multirow}
\usepackage{arydshln}
\usepackage{booktabs}
\usepackage{enumitem}
\usepackage{bm}

\definecolor{green}{rgb}{0.37,0.69,0.34}

\newcommand{\redbf}[1]{\textbf{\textcolor{red}{#1}}}
\newcommand{\rednm}[1]{\textcolor{red}{#1}}
\newcommand{\greennm}[1]{\textcolor{green}{#1}}

\usepackage{wrapfig}
\usepackage{fontawesome}

\def\ie{\emph{i.e.}}
\def\eg{\emph{e.g.}}

\def\etal{{\em et al.}}

\title{X$^2$Localizer: Cross-grained Alignment for Progressive Cross-view Video Geo-localization}

\addauthor{Zichao Zeng}{zichao.zeng.21@ucl.ac.uk}{1,2}
\addauthor{Weijia Fan}{weijia.fan@ualberta.ca}{2,4,5}
\addauthor{Yufan Chen}{yufan.chen@kit.edu}{2}
\addauthor{June Moh Goo}{june.goo.21@ucl.ac.uk}{1}
\addauthor{Junwei Zheng}{junwei.zheng@kit.edu}{2,$\dag$}
\addauthor{Ruiping Liu}{ruiping.liu@kit.edu}{2}
\addauthor{Kunyu Peng}{kunyu.peng@kit.edu}{2}
\addauthor{Jiaming Zhang}{jiamingzhang@hnu.edu.cn}{3,$\dag$}
\addauthor{Rainer Stiefelhagen}{rainer.stiefelhagen@kit.edu}{2}
\addauthor{Jan Boehm}{j.boehm@ucl.ac.uk}{1}

\addinstitution{
 University College London,\\
 London, UK
}
\addinstitution{
 Karlsruhe Institute of Technology,\\
 Karlsruhe, Germany
}
\addinstitution{
 Hunan University,\\
 Changsha, China
}
\addinstitution{
 University of Alberta,\\
 Edmonton, Canada
}
\addinstitution{
 Shenzhen University,\\
 Shenzhen, China
}
\runninghead{Zeng \emph{et al}\bmvaOneDot}{X$^2$Localizer}

\def\eg{\emph{e.g}\bmvaOneDot}

\def\etal{\emph{et al}\bmvaOneDot}

\begin{document}

\maketitle

\begin{abstract}
\sloppy
Cross-view Video Geo-localization (CVG) aims to localize ground-view videos by retrieving their corresponding geo-tagged aerial images. However, CVG approaches rely on fixed-length inputs and post-hoc refinement, hindering online-oriented localization under partial or dynamic observations. 
In this work, we formulate Progressive Cross-view Video Geo-localization (PCVG) as a deployment-oriented extension and evaluation protocol of CVG, enabling localization under varying temporal budgets, prefix-based inference, random-start evaluation, and long-range localization with interruptions.
To explore PCVG, we introduce X$^2$Localizer, a cross-grained alignment framework that jointly supervises global prefix-to-aerial retrieval and token-aggregated frame--aerial-tile matching with a budget-dependent asymmetric objective.
Furthermore, we introduce a Sliding-Window Re-Localization (SWRL) strategy that dynamically refreshes candidate regions for failure recovery and long-range deployment without full-sequence reprocessing.
Extensive experiments show that X$^2$Localizer preserves conventional full-video performance, with marginal gains of +0.1 Recall@1 and +0.3 Recall@10, while substantially improving early localization.
In the challenging single-frame setting, X$^2$Localizer improves coarse retrieval by +4.7 Recall@1 and +11.5 Recall@10 over the previous state-of-the-art method. 
With SWRL, our approach further enables robust progressive localization under random-start and long-distance scenarios, narrowing the gap between benchmark evaluation and real-world deployment. \\
\faCode~The code is publicly available at~\href{https://zichaozeng.github.io/X2Localizer}{https://zichaozeng.github.io/X2Localizer}.
\end{abstract}

\section{Introduction}
\label{sec:intro}
Visual geo-localization aims to estimate the geographic location of a query image by matching it against geo-tagged reference imagery~\cite{WeiToP,SelV,LLMCG,SD-MCG,TransGeoCG,StatewideCG,Sample4GeoCG,GeoDistillMCG,NoGroundTruthMCG}. Cross-view geo-localization, which aligns ground-view observations with aerial or satellite imagery, has attracted significant attention due to its applications in autonomous navigation, robotics, digital twins, and urban computing~\cite{CG4Urban1,CG4Urban2,DigitalTwins,DigitalTwins2,CG4SLAM,CG4Auto}. While early studies focus on single-image matching, recent works extend the problem to \textit{Cross-view Video Geo-localization (CVG)}, where a ground-view video is matched against a large aerial image~\cite{GAMa,GAReT,CVG1,SeqGeo,CVLNet}. By aggregating temporal cues across frames, these methods consistently outperform single-frame approaches, demonstrating the effectiveness of spatio-temporal modeling. 

Despite these advances, early methods in cross-view geo-localization either aggregate sequence features to predict a single coarse location for the entire video~\cite{SeqGeo}, or rely on explicit geometric projection based on camera parameters and estimated relative poses to align ground and aerial views~\cite{CVLNet}. More recent approaches achieve frame-level localization using purely visual representations~\cite{GAReT, GAMa}. However, they are typically designed for offline inference, assuming access to the complete query sequence before trajectory estimation, and do not explicitly consider streaming or incremental localization settings. Such an evaluation protocol does not align with real-world deployment scenarios. In practical systems, video streams arrive progressively, and localization must be performed incrementally~\cite{SeqSLAM,SeqSLAM2,SeqVPR}. The model should produce reliable predictions from short prefixes (\eg, a single frame or a few seconds), adapt to arbitrary starting timestamps, and remain robust to interruptions or partial observations.
Empirically, we observe that when evaluated under such progressive conditions, existing CVG methods exhibit noticeable performance degradation, revealing a fundamental gap between benchmark assumptions and practical requirements.

\sloppy To bridge this gap, we reformulate the task as \textit{Progressive Cross-view Video Geo-localization (PCVG)}. Instead of assuming access to a complete fixed-length sequence, we require the model to localize under varying temporal budgets, ranging from single-frame to full-length videos. To enable systematic evaluation, we reconstruct the protocol of the GAMa dataset~\cite{GAMa} and establish a new progressive benchmark that supports \textit{multi-duration prefix evaluation}, \textit{random-start testing}, and \textit{long-distance or interruption scenarios}. This reformulation not only provides a more realistic evaluation setting, but also exposes intrinsic limitations of existing coarse-grained global matching strategies. 

\begin{figure}[t]
    \centering
    \includegraphics[width=1.0\linewidth]{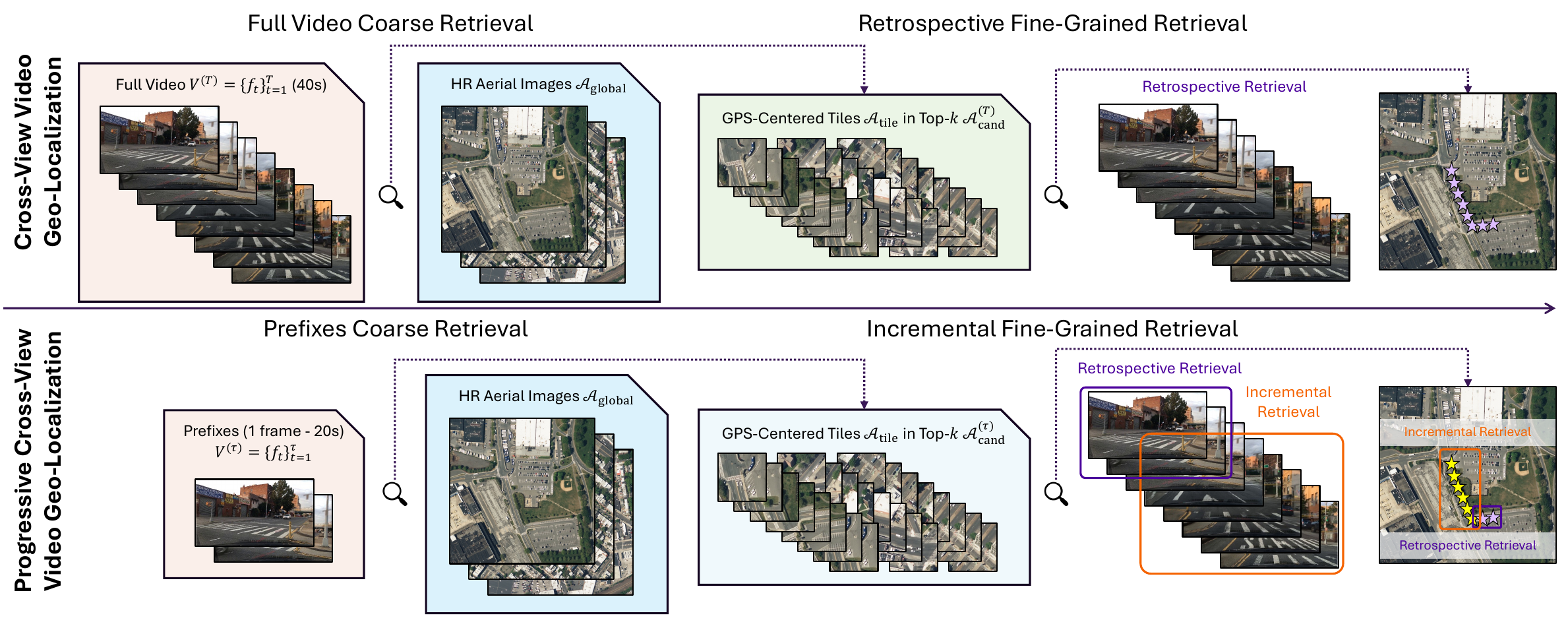}
    \caption{Compared to traditional cross-view video geo-localization (CVG), our proposed Progressive CVG (PCVG) enables precise localization (1) at arbitrary timestamps, (2) for varying video lengths, and (3) remains robust to interruptions or missing frames.}
    \label{fig:PCVGL_vs_CVGL}
\end{figure}

A central challenge in PCVG lies in the limited contextual information available in short video segments. 
Conventional approaches primarily rely on global-to-global matching between a full aerial image and a temporally aggregated ground-video representation.
Such a design is inherently fragile when only partial observations are available~\cite{SeqSLAM}.
To address this issue, we propose \textbf{X$^2$Localizer}, a \textbf{cross-grained} and \textbf{cross-view} alignment framework that extends existing CVG methods~\cite{GAReT,TransGeoCG} to the PCVG setting.
Inspired by multi-grained contrastive learning~\cite{CLIP,Filip,XCLIP,TACO}, our key insight is to establish multi-grained correspondences between global and local aerial representations and ground-level temporal observations under varying temporal budgets.
X$^2$Localizer therefore jointly models global image-to-video alignment and fine-grained patch-to-frame alignment, with an asymmetric objective that places greater emphasis on local frame-level cues for shorter prefixes and stronger global alignment for longer observations.
By modeling these cross-grained interactions within a unified contrastive learning framework, X$^2$Localizer enables reliable cross-view coarse localization under varying temporal budgets, naturally facilitating subsequent incremental frame-to-frame refinement and re-localization.
Furthermore, inspired by ground-level sequence-based localization methods~\cite{LiDARPR,MultiSeqVPR}, we introduce a \textbf{Sliding-Window Re-Localization (SWRL)} strategy that preserves stable alignment signals even in long-range or interrupted video streams.
Our contributions are summarized as follows:
\begin{itemize}[nosep]
    \item 
    We redefine cross-view video geo-localization as \textbf{Progressive Cross-view Video Geo-localization (PCVG)}, a deployment-oriented setting supporting multi-duration, random-start, and long-distance evaluation. We reconstruct the GAMa dataset protocol to establish a new progressive benchmark.
    \item 
    We propose \textbf{X$^2$Localizer}, a cross-grained alignment framework that combines global prefix-to-aerial alignment with token-aggregated frame–aerial-tile alignment. It asymmetrically weights these objectives according to the available temporal context, enabling robust localization under partial observations.
    \item 
    We introduce a \textbf{Sliding-Window Re-Localization (SWRL)} inference strategy that allows dynamic re-localization over long video streams, enabling failure recovery and sustained long-range deployment.
    \item 
    Extensive experiments demonstrate that \textbf{X$^2$Localizer} achieves competitive performance under the conventional full-video protocol, while substantially improving prefix and progressive localization performance under the proposed PCVG benchmark.
\end{itemize}

\section{Related Work}
\label{sec:related_work}

\paragraph{Cross-view Image Geo-localization.}
Cross-view image geo-localization aims to match ground-view images to geo-tagged aerial or satellite imagery~\cite{TransGeoCG,StatewideCG,Sample4GeoCG,RHO}. Early methods primarily rely on Siamese or triplet-based metric learning frameworks to learn view-invariant representations. Representative works include SAFA~\cite{SAFA}, DSM~\cite{DSM1,DSM2}, and L2LTR~\cite{L2LTR}, which introduce orientation alignment, polar transformation, or dynamic similarity matching to mitigate severe viewpoint discrepancies between ground and aerial views. Subsequent approaches explore attention mechanisms and transformer-based architectures to model global context more effectively. For example, TransGeo~\cite{TransGeoCG} demonstrates that pure transformer models can achieve strong cross-view alignment without explicit geometric transformations. More recent works further investigate fine-grained correspondence learning and pose-aware modeling to improve localization precision~\cite{CNNCPG,BeyondCPG,DenUncerCPG,SlicematchCPG,UncertantyCPG} or leverage the ability of large language models~\cite{LLMCG}. However, these methods operate on single images and do not model temporal continuity, making them insufficient for video-based progressive localization.

\paragraph{Cross-view Video Geo-localization.}
To mitigate the limited field-of-view of single ground images, SeqGeo~\cite{SeqGeo} aggregates short ground-view sequences for cross-view matching, demonstrating improved robustness over single-frame methods. Extending this direction, GAMa~\cite{GAMa} introduces the first large-scale cross-view video dataset with a hierarchical coarse-to-fine strategy, while CVLNet~\cite{CVLNet} incorporates geometric projection and temporal constraints but relies on camera intrinsics and odometry. More recently, GAReT~\cite{GAReT} adapts image geo-localization models to video via lightweight adapters and autoregressive retrieval, achieving state-of-the-art performance under fixed-length settings. Despite their strong fixed-length performance, existing CVG methods generally assume that the complete query video is available before inference. In contrast, ground-level sequence-based localization has highlighted the importance of progressive inference, incremental updates, and re-localization for real-world robustness~\cite{SeqSLAM,SeqSLAM2,LiDARPR,MultiSeqVPR,SeqVPR,SeqLoc}. 
PCVG transfers these operational requirements to cross-view video geo-localization, where frames arrive incrementally and the system must handle arbitrary starts, interruptions, and cross-region transitions.

\section{Methodology}
\label{sec:method}
\subsection{Problem Formulation}
\label{sec:problem}
The proposed task of \textbf{PCVG} is shown in Fig.~\ref{fig:PCVGL_vs_CVGL}. Given a ground-view video sequence and a geo-tagged aerial image database, the goal is to localize the video frames by cross-view retrieval under \textit{varying temporal budgets}.
Let $\mathcal{V} = \{V_i\}_{i=1}^{N}$ denote a set of ground-view videos.  Each video $V = \{\bm f_t\}_{t=1}^{T}$ consists of $T$ ordered frames. For each frame $\bm f_t$, there exists a corresponding geo-tagged aerial tile (small GPS-centered image) $\bm a_t \in \mathcal{A}_{\text{tile}}$. In addition, each video $V$ is associated with a high-resolution aerial image $A^{\text{global}} \in \mathcal{A}_{\text{global}}$ that covers the entire geographic region of the trajectory. Let $\mathcal{A}_{\text{global}}$ and $\mathcal{A}_{\text{tile}}$ denote the global aerial gallery and tile-level aerial gallery, respectively.

Existing CVG methods~\cite{GAMa,GAReT} assume access to the complete video $V$ before localization. They typically perform:
(1) coarse retrieval by matching the full video representation to $\mathcal{A}_{\text{global}}$, and 
(2) backtracking fine-grained frame-tile retrieval within the selected region by $\mathcal{A}_{\text{tile}}$. Formally, a ground-view encoder $\phi_v$ and an aerial tile encoder $\phi_a$ are pretrained by frame-tile matching, \ie, $\phi_v(\bm f_t) \approx \phi_a(\bm a_t)$. Subsequently, with lightweight adapters, a video encoder $\Phi_v$ and a global aerial image encoder $\Phi_a$ are learned such that $\Phi_v(V^{(T)}) \approx \Phi_a(A^{\text{global}})$. However, during inference, $\Phi_v$ and $\Phi_a$ are first employed for coarse retrieval, followed by $\phi_v$ and $\phi_a$ backtracking to retrieve each frame individually. This formulation implicitly assumes a fully observed and uninterrupted video sequence.

In contrast, we consider a \textit{progressive} formulation where a prefix of the video is observable. Let $V^{(\tau)} = \{\bm f_t\}_{t=1}^{\tau}$ denote the first $\tau$ frames of $V$, where $\tau \in \{1, \dots, T\}$. The model is required to perform localization under varying temporal budgets $\tau$, including single-frame ($\tau=1$), short-clip, half-length, and full-length ($\tau=T$) scenarios. 
Our objective is to learn representations that remain discriminative for every prefix length $\tau$. Specifically, the cosine similarity
$\cos\!\left(\Phi_v(V^{(\tau)}),\Phi_a(A^{\mathrm{global}})\right)$ should be maximized for the correct aerial candidate and suppressed for mismatched candidates.
Simultaneously, fine-grained frame–tile alignment is preserved through 
$\cos(\phi_v(\bm f_t), \phi_a(\bm a_t))$.
This progressive formulation introduces two key challenges: (i) limited contextual information when $\tau$ is small, and (ii) robustness under interruption, restart, or cross-region transitions in long videos. We refer to this deployment-oriented formulation as \textit{PCVG}.

\begin{figure*}[t]
  \centering

  \begin{minipage}[t]{0.39\textwidth}
    \centering
    \includegraphics[width=\linewidth]{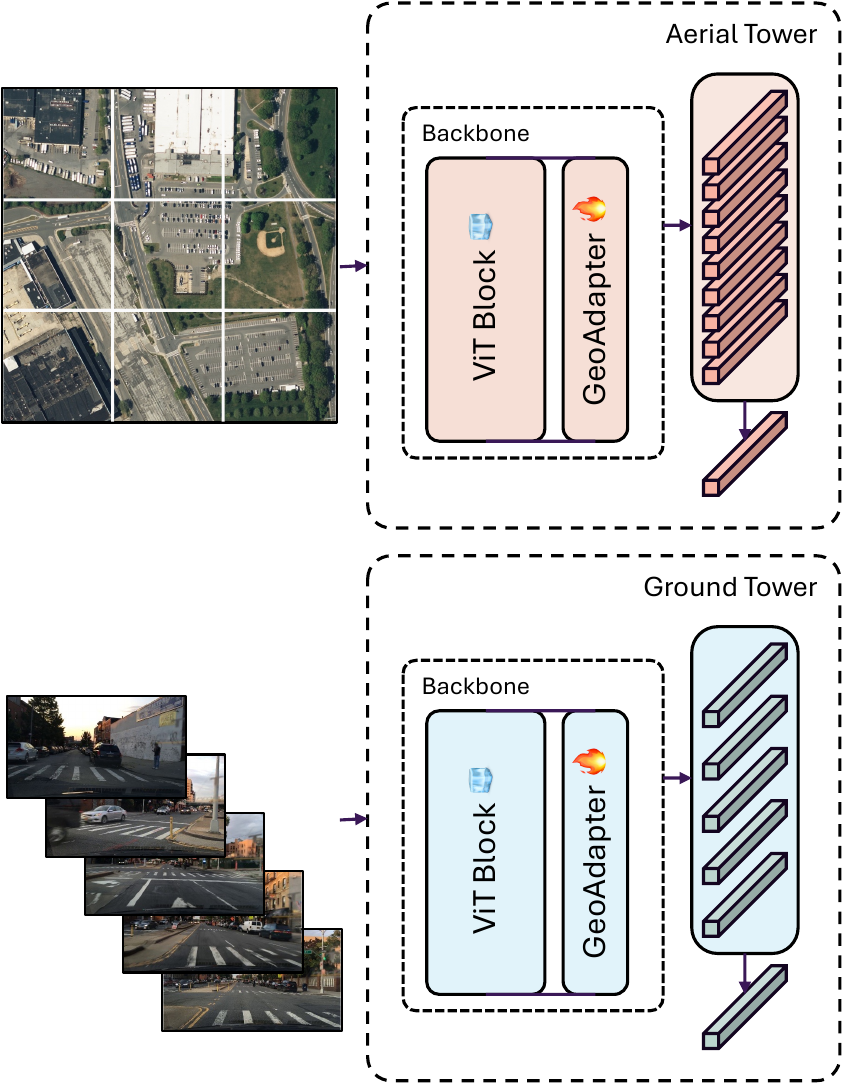}

    \small \textbf{(a)} Adaptation
    \vspace{1mm}
  \end{minipage}
  \hspace{0.01\textwidth}
  \begin{minipage}[t]{0.58
\textwidth}
    \centering
    \includegraphics[width=\linewidth]{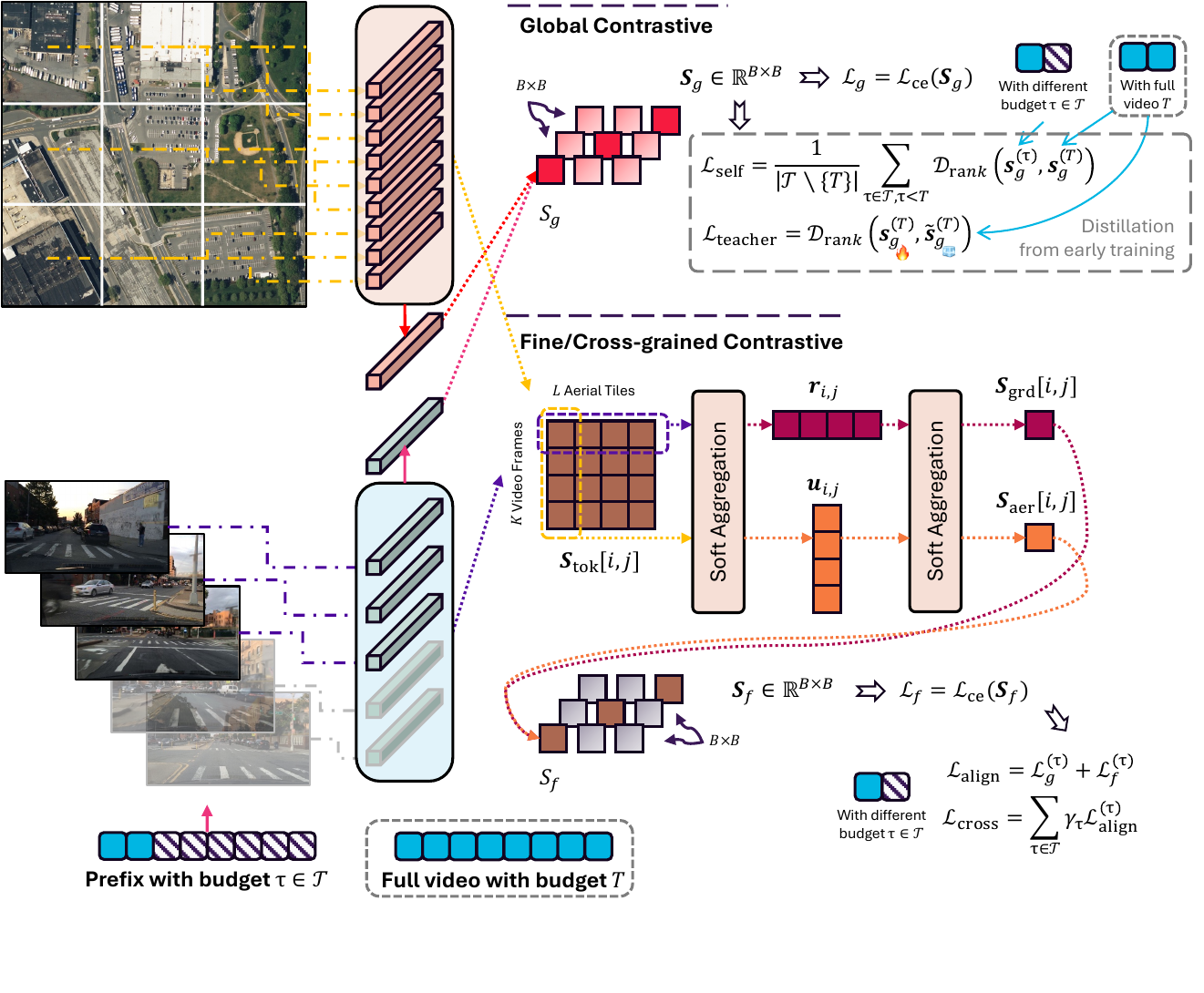}

    \small \textbf{(b)} Cross-grained Alignment Objective
    \vspace{1mm}
  \end{minipage}

  \caption{
  (a) GeoAdapter adapts frozen pretrained dual-tower encoders from image-tile matching to video-aerial matching. 
    (b) For each temporal budget, our objective combines global video-to-aerial alignment, frame-tile alignment, and ranking distillation; shorter prefixes receive stronger supervision, while longer prefixes emphasize global alignment.
  }
  \label{fig:methodology}
\end{figure*}

\subsection{Pretraining and Adaptation for Cross-view Representation}
\label{sec:pretrain_adapt}
Following previous works~\cite{TransGeoCG,GAReT,SAFA,GEODTR1,GEODTR2,VIGOR}, we adopt a dual-tower architecture with a ground-view encoder $\phi_v$ and an aerial-tile encoder $\phi_a$. 
Similar to~\cite{TransGeoCG,GAReT}, both encoders use the same distilled ViT backbone~\cite{DeiT} but do not share weights. 
Given an image input $\bm X$, the image representation is obtained by averaging the projected classification and distillation tokens, \ie,
\begin{equation}
\phi_\star(\bm X) =\operatorname{L2Norm}\left(
(\mathbf{h}_{\mathrm{cls}}^{(\star)} +
\mathbf{h}_{\mathrm{dist}}^{(\star)})/2 
\right),
\quad \star \in \{v,a\}.
\end{equation}
We pretrain the dual-tower encoders using frame-to-aerial-tile pairs $(\bm f_t, \bm a_t)$ with a single-direction soft-margin contrastive loss $\mathcal{L}_{\mathrm{smcl}}$ from the ground side to the aerial side. 
The full pretraining objective is provided in the supplementary material.

To extend the pretrained image encoders $\phi_v$ and $\phi_a$ to video-to-global matching, we follow~\cite{GAReT,STAdapter,AIM} and insert a lightweight GeoAdapter module into each Transformer block, while freezing the pretrained spatial backbone (see Fig.~\ref{fig:methodology}(a)).
Similar to GAReT~\cite{GAReT}, we first optimize the GeoAdapter using complete videos and their corresponding global aerial images. 
Given a mini-batch of $B$ matched video-aerial pairs $\{(V_i^{(T)}, A_i^{\mathrm{global}})\}_{i=1}^{B}$, we stack the full-video embeddings and global aerial embeddings as
\begin{equation}
\mathbf{V}^{(g,T)}
=
[\mathbf{v}_1^{(g,T)}, \dots, \mathbf{v}_B^{(g,T)}]^\top \in \mathbb{R}^{B \times d},
\qquad
\mathbf{A}^{(g)}=[\mathbf{a}_1^{(g)}, \dots, \mathbf{a}_B^{(g)}]^\top\in \mathbb{R}^{B \times d},
\end{equation}
where $\mathbf{v}_i^{(g,T)}=\Phi_v(V_i^{(T)})$,  
$\mathbf{a}_i^{(g)}=\Phi_a(A_i^{\mathrm{global}})$, and $d$ denotes the aligned feature dimension. All embeddings are $\ell_2$-normalized, so their dot products correspond to cosine similarity.
The batch-wise global similarity is computed as
\begin{equation}
\bm s_g^{(T)}
=\mathbf{V}^{(g,T)}\mathbf{A}^{(g)\top}\in \mathbb{R}^{B \times B},
\end{equation}
where $\bm s_g^{(T)}[i,j]$ denotes the similarity between the $i$-th ground-view video and the $j$-th global aerial image. 
We use a row-wise cross-entropy retrieval loss from the ground/video side to the aerial side:
\begin{equation}
\mathcal{L}_{\mathrm{ce}}(\bm s)
=-\frac{1}{B}\sum_{i=1}^{B}\log\frac{\exp(\bm s[i,i]/\tau_c)}{\sum_{j=1}^{B}\exp(\bm s[i,j]/\tau_c)},
\label{eq:row_ce}
\end{equation}
where $\tau_c$ is the temperature parameter. 
The full-video adaptation objective is therefore 
$\mathcal{L}_{\mathrm{full}}=\mathcal{L}_{\mathrm{ce}}(\bm s_g^{(T)})$.
This objective adapts the cross-view representation to full-video global localization before introducing cross-grained alignment for different temporal budgets.

\subsection{Asymmetric Cross-grained Alignment Objective}
\label{sec:cross_grained}

In the early stage, we train the model only with the full-video adaptation objective in Sec.~\ref{sec:pretrain_adapt}, which provides stable global alignment between complete ground-view videos and global aerial images.
However, in PCVG, the model must align ground-view observations of different temporal lengths $\tau\in\mathcal{T}\subseteq\{1,\dots,T\}$ with the same global aerial gallery. 
Unlike previous CVG methods that mainly rely on full-video global supervision, we introduce an \textbf{asymmetric cross-grained alignment} objective to jointly supervise global video-aerial matching and fine-grained token-level (frame-tile) matching under different temporal budgets (Fig.~\ref{fig:methodology}(b)).

Given a temporal budget $\tau$, the prefix of the $i$-th video is denoted as $V_i^{(\tau)}=\{\bm f_{i,t}\}_{t=1}^{\tau}$. 
The adapted encoders produce the prefix-level video embedding $\mathbf{v}_i^{(g,\tau)}=\Phi_v(V_i^{(\tau)})\in\mathbb{R}^{d}$ and the global aerial embedding $\mathbf{a}_j^{(g)}=\Phi_a(A_j^{\mathrm{global}})\in\mathbb{R}^{d}$. 
For a mini-batch of $B$ matched video-aerial pairs, we stack the embeddings as $\mathbf{V}^{(g,\tau)}$ and $\mathbf{A}^{(g)}$.
The global prefix-to-aerial similarity is computed as
$\bm s_g^{(\tau)}=\mathbf{V}^{(g,\tau)}\mathbf{A}^{(g)\top}\in\mathbb{R}^{B\times B}$,
where $\bm s_g^{(\tau)}[i,j]$ measures the global similarity between the $i$-th video prefix and the $j$-th aerial image.
The global alignment loss is then defined as
$\mathcal{L}_{g}^{(\tau)}=\mathcal{L}_{\mathrm{ce}}(\bm s_g^{(\tau)})$, where $\mathcal{L}_{\mathrm{ce}}$ is the row-wise retrieval cross-entropy in Eq.~\ref{eq:row_ce}.

To provide fine-grained cross-view supervision, we further compute local token-level frame-tile similarities.
Let $\mathbf{P}_i^{(\tau)}=[\mathbf{p}_{i,1},\dots,\mathbf{p}_{i,K}]^\top\in\mathbb{R}^{K\times d}$ denote the $K$ ground/video tokens of the $i$-th prefix $V_i^{(\tau)}$, and let $\mathbf{Q}_j=[\mathbf{q}_{j,1},\dots,\mathbf{q}_{j,L}]^\top\in\mathbb{R}^{L\times d}$ denote the $L$ aerial tile tokens of the $j$-th global aerial image $A_j^{\mathrm{global}}$.
For each pair $(i,j)$, the token-token similarity is
\begin{equation}
\bm s_{\mathrm{tok}}^{(\tau)}[i,j,k,l]
=
\mathbf{p}_{i,k}^{\top}\mathbf{q}_{j,l},
\qquad
\bm s_{\mathrm{tok}}^{(\tau)}
\in
\mathbb{R}^{B\times B\times K\times L}.
\end{equation}
where $K=\tau$ in $V_i^{(\tau)}$ and $L$ is the number of aerial tiles from $A_j^{\mathrm{global}}$. Thus, each video-aerial pair has a $K\times L$ frame-tile similarity map, while the whole mini-batch forms a four-dimensional similarity tensor.

We aggregate the token-level similarities with a two-stage soft aggregation. 
For each ground/video token, we first softly aggregate over aerial tile tokens:
\begin{equation}
r_{i,j,k}^{(\tau)}
=
\sum_{l=1}^{L}
\alpha_{i,j,k,l}^{(\tau)}
\bm s_{\mathrm{tok}}^{(\tau)}[i,j,k,l],
\qquad
\alpha_{i,j,k,l}^{(\tau)}
=
\frac{
\exp(\bm s_{\mathrm{tok}}^{(\tau)}[i,j,k,l]/\tau_f)
}{
\sum_{l'=1}^{L}
\exp(\bm s_{\mathrm{tok}}^{(\tau)}[i,j,k,l']/\tau_f)
}.
\end{equation}
Here, $\tau_f$ is the token soft-aggregation temperature. We then softly aggregate the resulting ground-token scores:
\begin{equation}
\bm s_{\mathrm{grd}}^{(\tau)}[i,j]
=
\sum_{k=1}^{K}
\rho_{i,j,k}^{(\tau)}
r_{i,j,k}^{(\tau)},
\qquad
\rho_{i,j,k}^{(\tau)}
=
\frac{
\exp(r_{i,j,k}^{(\tau)}/\tau_f)
}{
\sum_{k'=1}^{K}
\exp(r_{i,j,k'}^{(\tau)}/\tau_f)
}.
\end{equation}
Symmetrically, for each aerial tile token, we softly aggregate over ground/video tokens:
\begin{equation}
u_{i,j,l}^{(\tau)}
=
\sum_{k=1}^{K}
\beta_{i,j,k,l}^{(\tau)}
\bm s_{\mathrm{tok}}^{(\tau)}[i,j,k,l],
\qquad
\beta_{i,j,k,l}^{(\tau)}
=
\frac{
\exp(\bm s_{\mathrm{tok}}^{(\tau)}[i,j,k,l]/\tau_f)
}{
\sum_{k'=1}^{K}
\exp(\bm s_{\mathrm{tok}}^{(\tau)}[i,j,k',l]/\tau_f)
},
\end{equation}
followed by a soft aggregation over aerial tokens:
\begin{equation}
\bm s_{\mathrm{aer}}^{(\tau)}[i,j]
=
\sum_{l=1}^{L}
\omega_{i,j,l}^{(\tau)}
u_{i,j,l}^{(\tau)},
\qquad
\omega_{i,j,l}^{(\tau)}
=
\frac{
\exp(u_{i,j,l}^{(\tau)}/\tau_f)
}{
\sum_{l'=1}^{L}
\exp(u_{i,j,l'}^{(\tau)}/\tau_f)
}.
\end{equation}
The final fine-grained similarity is
\begin{equation}
\bm s_f^{(\tau)}[i,j]
=
\frac{1}{2}
\left(
\bm s_{\mathrm{grd}}^{(\tau)}[i,j]
+
\bm s_{\mathrm{aer}}^{(\tau)}[i,j]
\right),
\qquad
\bm s_f^{(\tau)}
\in
\mathbb{R}^{B\times B}.
\end{equation}
Although $\bm s_f^{(\tau)}$ is obtained by aggregating token similarities from two complementary directions, the retrieval loss is applied only in the ground-to-aerial direction:
$\mathcal{L}_{f}^{(\tau)}=\mathcal{L}_{\mathrm{ce}}(\bm s_f^{(\tau)})$.


The asymmetric cross-grained alignment objective for temporal budget $\tau$ is
\begin{equation}
\mathcal{L}_{\mathrm{align}}^{(\tau)} = \lambda_{g}^{(\tau)} \mathcal{L}_{g}^{(\tau)} + \lambda_{f}^{(\tau)} \mathcal{L}_{f}^{(\tau)}.
\end{equation}
The weights $\lambda_{g}^{(\tau)}$ and $\lambda_{f}^{(\tau)}$ balance global and fine-grained supervision for different temporal lengths. 
Shorter prefixes rely more on fine-grained local evidence, while longer prefixes and full videos place more emphasis on global video-aerial alignment.

In addition to the supervised alignment objective, we preserve the ranking structure learned from full-video adaptation through row-wise ranking distillation.
Given a student similarity matrix $\bm s$ and a teacher similarity matrix $\tilde{\bm s}$, we define
\begin{equation}
\mathcal{D}_{\mathrm{rank}}(\bm s,\tilde{\bm s})
=
\tau_d^2
\operatorname{KL}
\left(
\operatorname{softmax}(\tilde{\bm s}/\tau_d)
\;\middle\|\;
\operatorname{softmax}(\bm s/\tau_d)
\right),
\end{equation}
where $\tau_d$ is the distillation temperature, and the softmax and KL divergence are computed row-wise over the aerial gallery.
For prefix-to-full self-distillation, shorter prefixes are encouraged to match the full-prefix ranking distribution produced by the current student:
\begin{equation}
\mathcal{L}_{\mathrm{self}}
=
\frac{1}{|\mathcal{T}\setminus\{T\}|}
\sum_{\tau\in\mathcal{T},\tau<T}
\mathcal{D}_{\mathrm{rank}}
\left(
\bm s_g^{(\tau)},
\bm s_g^{(T)}
\right).
\end{equation}
We further use the frozen full-video model from the early training stage as a teacher for the full-prefix student:
\begin{equation}
\mathcal{L}_{\mathrm{teacher}}
=
\mathcal{D}_{\mathrm{rank}}
\left(
\bm s_g^{(T)},
\tilde{\bm s}_g^{(T)}
\right),
\end{equation}
where $\tilde{\bm s}_g^{(T)}$ denotes the global similarity matrix produced by the early stage full-video teacher.

The final training objective is
\begin{equation}
\mathcal{L}_{\mathrm{cross}}
=
\sum_{\tau\in\mathcal{T}}
\gamma_{\tau}
\mathcal{L}_{\mathrm{align}}^{(\tau)},
\qquad
\mathcal{L}_{\mathrm{total}}
=
\mathcal{L}_{\mathrm{cross}}
+
\eta_{\mathrm{self}}
\mathcal{L}_{\mathrm{self}}
+
\eta_{\mathrm{teacher}}
\mathcal{L}_{\mathrm{teacher}}.
\label{eq:total_loss}
\end{equation}
Here, $\gamma_\tau$ weights temporal budget $\tau$, while $\eta_{\rm self}$ and $\eta_{\rm teacher}$ weight the two distillation terms.

\begin{figure*}[t]
  \centering

  \begin{minipage}[t]{0.40\textwidth}
    \centering
    \includegraphics[width=\linewidth]{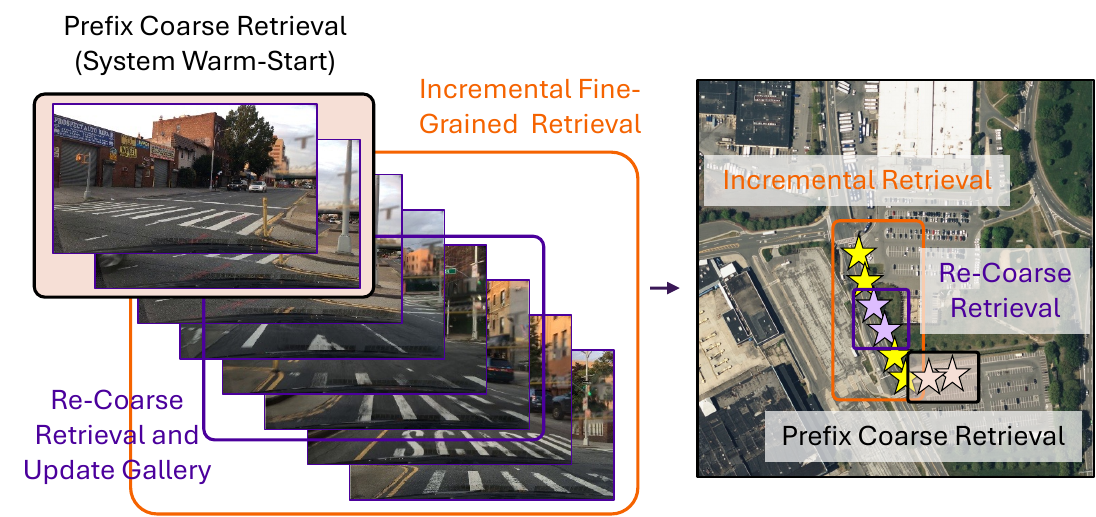}

    \small \textbf{(a)} SWRL Strategy
    \vspace{1mm}
  \end{minipage}
  \hspace{0.01\textwidth}
  \begin{minipage}[t]{0.53\textwidth}
    \centering
    \includegraphics[width=\linewidth]{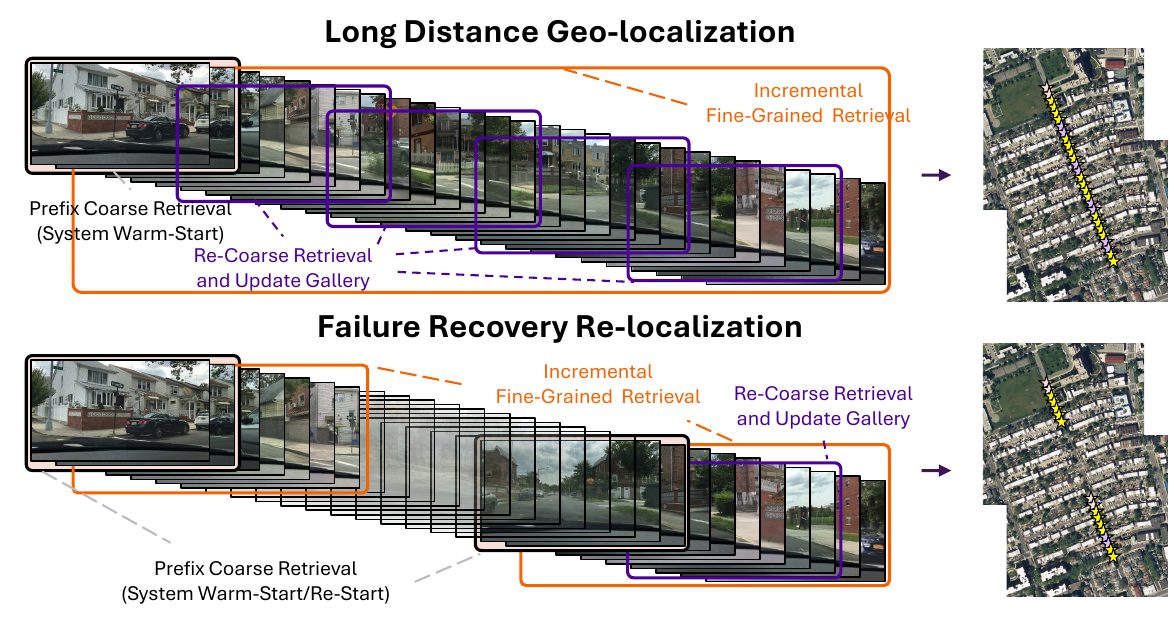}

    \small \textbf{(b)} Challenging Cases
    \vspace{1mm}
  \end{minipage}

  \caption{
    (a) Sliding-window Re-localization (SWRL) refreshes the candidate region set $\mathcal{A}_{\text{cand}}$ every $\Delta$ frames to support subsequent incremental refinement.
    (b) It improves robustness in long-range localization and enables failure recovery.
  }
  \label{fig:SWRL}
\end{figure*}

\subsection{Progressive Inference Strategy}
\label{sec:inference}

During inference, we localize a query video with either a prefix or the full sequence. 
Given $V^{(\tau)}=\{\bm f_t\}_{t=1}^{\tau}$, where $\tau\in\{1,\dots,T\}$, we compute $\mathbf{v}^{(g,\tau)}=\Phi_v(V^{(\tau)})$. 
For each global aerial candidate $A_j^{\mathrm{global}}\in\mathcal{A}_{\mathrm{global}}$, we use the same mixed-resolution matching strategy as in training:
\begin{equation}
s_{\mathrm{mix}}^{(\tau)}(j)
=
\frac{1}{2}
\left(
s_g^{(\tau)}(j)
+
s_f^{(\tau)}(j)
\right),
\qquad
s_g^{(\tau)}(j)
=
\mathbf{v}^{(g,\tau)\top}\mathbf{a}_j^{(g)}.
\label{eq:coarse}
\end{equation}
Here, $s_f^{(\tau)}(j)$ is obtained by applying the same two-direction token aggregation to the frame-token and aerial-tile-token similarities. 
We select the top-$K_c$ global aerial regions according to $s_{\mathrm{mix}}^{(\tau)}(j)$ to form $\mathcal{A}_{\mathrm{cand}}^{(\tau)}$. 
When $\tau=T$, this becomes full-video retrieval; when $\tau<T$, it enables early localization with partial observations.

For long video streams that may traverse multiple geographic regions, an initially retrieved candidate set may gradually become suboptimal. 
Inspired by~\cite{MultiSeqVPR,LiDARPR}, we introduce the \textbf{Sliding-window Re-localization (SWRL)} strategy to periodically refresh the candidate set (Fig.~\ref{fig:SWRL}(a)). 
Given a continuous stream $\{\bm f_t\}_{t=1}^{\infty}$, we define a sliding window $W^{(t)}$. Within this window, we construct a window-prefix $W^{(t,\tau)}$, which is fed into the video encoder.
\begin{equation}
W^{(t)}
=
\{\bm f_{t-\Delta+1},\dots,\bm f_t\}, \qquad
W^{(t,\tau)}
=
\{\bm f_{t-\Delta+1},\dots,\bm f_{t-\Delta+\tau}\},
\qquad
\tau\in\{1,\dots,\Delta\},
\end{equation}
where $\Delta$ denotes the window size used for re-localization.
The window-prefix embedding and its mixed-resolution retrieval score are computed as
\begin{equation}
\mathbf{v}_{W}^{(g,\tau)}
=
\Phi_v(W^{(t,\tau)}),
\qquad
s_{W}^{(\tau)}(j)
=
s_{\mathrm{mix}}^{(\tau)}(W^{(t,\tau)},A_j^{\mathrm{global}}),
\end{equation}
where $s_{\mathrm{mix}}^{(\tau)}$ is defined in Eq.~\ref{eq:coarse}.

The first coarse retrieval requires accumulating a prefix, which introduces a warm-start cost before localization is initialized. 
After initialization, frame-level geo-localization is performed using the candidate tile gallery. 
SWRL periodically refreshes the candidate set using the prefix of the current sliding window $W^{(t,\tau)}$, 
and the geo-tagged tile reference gallery is updated. 
This mechanism enables (1) \emph{cross-region or long-distance localization} when the trajectory moves beyond the initially retrieved geographic area, and (2) \emph{failure recovery} under occlusion, visual ambiguity, or temporary interruption, by reinitializing localization using only a short temporal budget within the latest window (Fig.~\ref{fig:SWRL}). 
SWRL requires no additional training or supervision and naturally integrates with progressive inference.

\section{Experiments and Results}
\label{sec:experiment}
\subsection{Implementation Details}
\label{sec:implementation}

We adopt DeiT-Small~\cite{DeiT}, pretrained on ImageNet~\cite{ImageNet}, as the backbone for ground-view and aerial branches in the dual-tower architecture. The image-level encoders $\phi_v$ and $\phi_a$ are pretrained with frame--aerial-tile pairs using the ground-to-aerial soft-margin contrastive objective. We then freeze the pretrained spatial backbone and insert GeoAdapter modules into the Transformer blocks to obtain the video/global-aerial encoders $\Phi_v$ and $\Phi_a$. The early adaptation stage uses only the full-video global retrieval loss $\mathcal{L}_{\mathrm{full}}$ to train the adapters, and is run for at most 50 epochs with early stopping patience 10; the resulting full-video model is used as the teacher for progressive training. In the progressive stage, we train with temporal budgets $\tau\in\mathcal{T}=\{1,2,4,8\}$, with $K=\tau$, corresponding to the first frame, 5s, 20s, and the full 40s video. For $\tau=1,2,4,8$, respectively, the budget weights $\gamma_\tau$ are set to $(0.05,0.10,0.25,2.00)$. We set the asymmetric component weights $(\lambda_g^{(\tau)},\lambda_f^{(\tau)})$ to $(1,2)$, $(1,1)$, $(1,0.5)$, and $(1,0)$, so short prefixes receive stronger fine-grained supervision while the full video is supervised only by global alignment. Prefix-to-full self-distillation and early-teacher ranking distillation are weighted by $\eta_{\rm self}=0.2$ and $\eta_{\rm teacher}=1.0$, respectively. We set the contrastive and distillation temperatures $\tau_c=\tau_d=0.07$ and the token soft-aggregation temperature $\tau_f=0.01$. Both adaptation stages use Adam with learning rate $1\times10^{-4}$, batch size $B=8$, mixed precision, and are trained for at most 50 epochs with patience 10 on a single NVIDIA RTX PRO 6000 GPU. For SWRL, we refresh candidate regions every 20 seconds ($\Delta$).


\paragraph{Dataset.}
Following prior work, we use the train-day split of the GAMa dataset for training and the val-day split for evaluation. GAMa is a cross-view geo-localization benchmark designed for frame-to-frame matching. Each sample contains a $\sim$40s street-view video (from BDD100K) paired with (1) a global aerial image covering the surrounding region, and (2) frame-level geo-tagged aerial image tiles corresponding to each video frame. The train-day split contains 21,144 video-global aerial pairs and approximately 790K frame-level aerial matches. The val-day split includes 3,103 videos and around 116K frame-level matches. To evaluate long-range localization ability, we further construct a \textit{val-long-distance} subset by concatenating two sequences from the same geographic source whose temporal gap is no more than two minutes. This subset contains 127 long sequences and approximately 6K frame-level pairs for evaluation.

\paragraph{Evaluation Protocols.}
We evaluate X$^2$Localizer under both the conventional CVG protocol and the proposed PCVG protocol. The conventional protocol assumes that the full 40-second video is available before coarse retrieval and retrospective frame-level localization, whereas PCVG evaluates localization under partial, restarted, and long-range streaming observations. We consider four settings. First, for global coarse retrieval, we retrieve the corresponding global aerial image using $\mathcal{T}=\{1,2,4,8\}$, covering both shorter prefixes and the full video. Second, for coarse-to-fine frame-level localization, the prefix-level coarse retriever selects the top-10 global aerial candidate regions, from which we construct the tile gallery and rerank frame-level aerial tiles using the image-level retriever. A frame prediction is correct if its GPS location is within 80 meters of the ground truth. Third, for random-start recovery, we sample one valid starting timestamp for each validation video, perform prefix-based coarse re-localization from that timestamp, and incrementally retrieve frame-level matches on the remaining frames. This setting simulates localization restart after interruption, tracking failure, or missing context. Fourth, for long-distance progressive localization, we evaluate continuous frame-level retrieval on the long-distance subset, where SWRL periodically refreshes candidate regions using the prefix of the current sliding window rather than relying on a single initial coarse retrieval. We report Recall@1, Recall@5, Recall@10, and Recall@1\%.

\subsection{Coarse Retrieval ($\mathcal{V}$-to-$\mathcal{A}_{\text{global}}$)}

We compare X$^2$Localizer with CVLNet~\cite{CVLNet}, GAMa~\cite{GAMa}, GAReT~\cite{GAReT}, and general-purpose video backbones including VideoSWIN~\cite{VideoSWIN} and TimeSformer~\cite{TimeSformer}. 
GAReT shares the same pretrained dual-tower cross-view encoder as our method but optimizes video-to-global matching mainly under the full-video setting. 
We also include two DeiT-based baselines to isolate the effect of the proposed objective. 
DeiT denotes a fine-tuned DeiT dual-tower model without GeoAdapter adaptation, while DeiT$^\star$ uses the same backbone but is trained with our asymmetric cross-grained alignment objective. 
Detailed baseline descriptions are provided in the supplementary material.


Table~\ref{tab:v2lai} reports coarse retrieval under different temporal budgets.
Under the conventional full-video protocol, X$^2$Localizer achieves performance comparable to GAReT, indicating that progressive training does not sacrifice standard CVG performance. 
The advantage becomes more evident under the PCVG protocol. 
When the input is shortened to 20 seconds, 5 seconds, or a single frame, X$^2$Localizer consistently improves over GAReT, with the largest gain appearing in the most constrained single-frame setting. 
This confirms that relying only on full-video global alignment is insufficient for progressive localization. 
By contrast, our asymmetric cross-grained objective directly supervises both prefix-level global retrieval and token-aggregated frame-tile evidence, allowing the representation to remain discriminative even when temporal context is limited.
In addition, comparing DeiT$^\star$ with DeiT shows that the proposed asymmetric cross-grained objective consistently improves retrieval even without GeoAdapter adaptation. 
For $\tau=\{1,2,4,8\}$, it improves Recall@1 by +4.9, +6.1, +7.1, and +6.3 for DeiT, respectively. 
This indicates that asymmetric cross-grained supervision is broadly beneficial for progressive retrieval, not only for the final X$^2$Localizer architecture.


\begin{table}[t]
\centering
\resizebox{0.8\columnwidth}{!}{%
\begin{tabular}{lccccccc}
\toprule[1.5pt]
\textbf{Model} & \textbf{Backbone} & \textbf{\#params} & \textbf{Latency} & \textbf{R@1} & \textbf{R@5} & \textbf{R@10} & \textbf{R@1\%}  \\ \midrule
\rowcolor{gray!15}\multicolumn{8}{l}{\textit{Full 40s video-to-global aerial image} $\tau=8$} \\
TimeSFormer~\cite{TimeSformer} & ViT-B & 243M & 14.2 & 20.1 & 44.5 & 55.6 & 83.5  \\ 
VideoSWIN~\cite{VideoSWIN} & Swin-B & 175M & 14.3 & 20.4 & 45.9 & 59.9 & 88.0  \\ 
CVLNet~\cite{CVLNet} & VGG16 & 17M & 23.9 & 0.4 & 1.3 & 2.7 & 15.4  \\ 
GAMa~\cite{GAMa} & Mixed & 23M & 32.7 & 12.2 & - & 35.3 & 49.3 \\ 
DeiT~\cite{DeiT} & DeiT-S & 45M & 2.9 & 20.5 & 49.9 & 63.6 & 83.2 \\ 
DeiT$^\star$~\cite{DeiT} & DeiT-S & 45M & 2.9 & 26.8 & 61.6 & 74.2 & 89.3 \\ 
GAReT~\cite{GAReT} & DeiT-S & 47M & 6.9 & 50.2 & 83.3 & 90.7 & 96.5 \\ \hdashline
\textbf{X$^2$Localizer (ours)} & DeiT-S & 47M & 6.9 & \textbf{50.3} & \textbf{83.9} & \textbf{91.0} & \textbf{97.7} \\ 
\qquad \textit{Improv.} & & & & \greennm{+0.1} & \greennm{+0.6} & \greennm{+0.3} & \greennm{+1.2} \\ 
  \midrule
\rowcolor{gray!15}\multicolumn{8}{l}{\textit{First 20s clip-to-global aerial image} $\tau=4$} \\
DeiT~\cite{DeiT} & DeiT-S & 45M & 2.1 & 16.2 & 43.0 & 57.7 & 78.1 \\ 
DeiT$^\star$~\cite{DeiT} & DeiT-S & 45M & 2.1 & 23.3 & 54.1 &  67.4 & 84.5 \\ 
GAReT~\cite{GAReT}   & DeiT-S & 47M & 5.2 & 41.4 & 75.0 & 84.4 & 93.6 \\  \hdashline
\textbf{X$^2$Localizer (ours)} & DeiT-S & 47M & 5.2 & \textbf{42.5} & \textbf{78.2} & \textbf{86.7} & \textbf{95.1}\\ 
\qquad \textit{Improv.} & & & & \greennm{+1.1} & \greennm{+3.2} & \greennm{+2.3} & \greennm{+1.5} \\ 
  \midrule
\rowcolor{gray!15}\multicolumn{8}{l}{\textit{First 5s clip-to-global aerial image} $\tau=2$} \\
DeiT~\cite{DeiT} & DeiT-S & 45M & 1.2 & 10.3 & 31.9 & 44.4 & 65.1 \\ 
DeiT$^\star$~\cite{DeiT} & DeiT-S & 45M & 1.2 & 16.4 & 42.7 & 56.2 & 75.6 \\ 
GAReT~\cite{GAReT}   & DeiT-S & 47M & 4.7 & 25.9 & 55.0 & 66.8 & 82.1 \\  \hdashline
\textbf{X$^2$Localizer (ours)} & DeiT-S & 47M & 4.7 & \textbf{29.1} & \textbf{61.9} & \textbf{72.8} & \textbf{86.9}\\ 
\qquad \textit{Improv.} & & & & \greennm{+3.2} & \greennm{+6.9} & \greennm{+6.0} & \greennm{+4.8} \\ 
  \midrule
\rowcolor{gray!15}\multicolumn{8}{l}{\textit{First frame-to-global aerial image} $\tau=1$} \\
DeiT~\cite{DeiT} & DeiT-S & 45M & 1.0 & 7.4 & 24.6 & 35.8 & 57.0 \\ 
DeiT$^\star$~\cite{DeiT} & DeiT-S & 45M & 1.0 & 12.3 & 35.6 & 48.6 & 67.4  \\ 
GAReT~\cite{GAReT}   & DeiT-S & 47M & 4.4 & 16.9 & 40.4 & 52.2 & 69.7 \\  \hdashline
\textbf{X$^2$Localizer (ours)} & DeiT-S & 47M & 4.4 & \textbf{21.6} & \textbf{50.9} & \textbf{63.7} & \textbf{80.7}  \\ 
\qquad \textit{Improv.} & & & & \greennm{+4.7} & \greennm{+10.5} & \greennm{+11.5} & \greennm{+11.0} \\ 
\addlinespace[0.5ex]
\bottomrule[1.5pt]
\end{tabular}
}
\vspace{1mm}
\caption{Coarse retrieval performance under varying prefix budgets $\tau$. Matching is evaluated between $V^{(\tau)}$ and $\mathcal{A}_{\mathrm{global}}$, reported in Recall@$k$ (\%) and inference latency (ms/video). Recall@1\% denotes top 31 of the full gallery. The best results are highlighted in \textbf{bold}.}\label{tab:v2lai}
\end{table}

\subsection{Fine-grained Retrieval ($\mathcal{V}$-to-$\mathcal{A}_{\text{tile}}$)}
We next evaluate whether better prefix-level coarse retrieval leads to stronger frame-level localization. 
Following prior work, a frame-level prediction is considered correct if the predicted GPS location falls within 0.05 miles (nearly 80 m) of the ground-truth location. 
We compare our method with \citet{DSM1}, L2LTR~\cite{L2LTR}, GAMa~\cite{GAMa} including its hierarchical variant GAMa$^\star$, and GAReT. 
For GAReT and X$^2$Localizer, the fine-grained tile gallery is constructed from the top-10 global aerial candidates retrieved by the corresponding coarse model.
Table~\ref{tab:f2f} reports frame-level localization under different coarse-retrieval budgets. 
In the full-video setting, X$^2$Localizer achieves performance comparable to GAReT, showing that the proposed progressive objective preserves the standard retrospective localization ability. 
Under shorter prefix budgets, however, X$^2$Localizer consistently improves fine-grained retrieval. 
The improvement is especially clear for the 5-second and single-frame settings, where the initial coarse gallery is more difficult to construct reliably. 
This demonstrates that the asymmetric cross-grained objective improves not only global aerial retrieval but also the quality of the downstream tile gallery.

We further evaluate SWRL in the same setting. 
Instead of relying on a single initial coarse retrieval, SWRL periodically refreshes the candidate aerial region using the latest sliding-window prefix. 
This dynamic update mitigates error accumulation and allows the fine-grained retriever to recover from suboptimal initial candidates. 
The gains are most pronounced under short warm-start budgets, confirming that progressive coarse re-localization is beneficial for practical online deployment.
However, broad-rank metrics may remain comparable or slightly decrease because the refreshed gallery focuses on the current local region.

\begin{table}[t]
\centering
\resizebox{0.8\columnwidth}{!}{
\setlength{\tabcolsep}{8pt} 
\begin{tabular}{lccccccc}
\toprule[1.5pt]
\textbf{Model} & \textbf{Backbone} & \textbf{\#params} & \textbf{Latency} & \textbf{R@1} & \textbf{R@5} & \textbf{R@10 } & \textbf{R@1\%}  \\ \midrule
\rowcolor{gray!15}\multicolumn{8}{l}{\textit{Full 40s video-to-global aerial image} $\tau=8$} \\
Shi \etal~\cite{DSM1} & VGG16 & 18M & 2.0 & 9.6 & 18.1 & 26.6 & 71.9 \\ 
L2LTR~\cite{L2LTR} & ViT-B & 196M & 12.7 & 11.7 & 20.8 & 28.2 & 87.1 \\ 
GAMa~\cite{GAMa} & Mixed & 49M & 4.3 & 15.2 & 27.2 & 33.8 & \textbf{91.9} \\   
GAMa$^\star$~\cite{GAMa} & Mixed & 72M & 41.3 & 18.3 & 27.6 & 32.7 & - \\ 
GAReT~\cite{GAReT} & DeiT-S & 45M & 3.6 & \textbf{46.8} & 71.8 & 81.1 & 91.4 \\ \hdashline
\textbf{X$^2$Localizer (ours)} & DeiT-S & 45M & 3.6 & 46.5 & \textbf{72.0}& \textbf{81.4} & 91.8 \\ \midrule
\rowcolor{gray!15}\multicolumn{8}{l}{\textit{First 20s clip-to-global aerial image} $\tau=4$} \\
GAReT~\cite{GAReT} & DeiT-S & 45M & 3.6 & 45.9 & 68.9 & 77.0 & 85.8 \\ 
\qquad + SWRL & & & 4.9(/20s) &\rednm{45.9} & 67.0 & 74.0 & 78.6 \\ \hdashline
\textbf{X$^2$Localizer (ours)} & DeiT-S & 45M & 3.6 & \textbf{46.3} & \textbf{70.3} & \textbf{78.8} & \textbf{87.7}  \\ 
\qquad + SWRL & & & 4.9(/20s) & \redbf{47.4} & 70.0 & 77.3 & 82.1 \\ \midrule
\rowcolor{gray!15}\multicolumn{8}{l}{\textit{First 5s clip-to-global aerial image} $\tau=2$} \\
GAReT~\cite{GAReT} & DeiT-S & 45M & 3.6 & 39.2 & 57.6 & 63.8 & 70.2 \\ 
\qquad + SWRL  & & & 4.9(/20s) &\rednm{44.8} & \rednm{62.5} & \rednm{66.4} & 67.8 \\ \hdashline
\textbf{X$^2$Localizer (ours)} & DeiT-S & 45M & 3.6 & \textbf{41.7} & \textbf{61.5} & \textbf{68.0} & \textbf{74.7} \\ 
\qquad + SWRL & & & 4.9(/20s) & \redbf{48.0} & \redbf{67.3} & \redbf{71.6} & 72.9 \\ \midrule
\rowcolor{gray!15}\multicolumn{8}{l}{\textit{First frame-to-global aerial image} $\tau=1$} \\
GAReT~\cite{GAReT} & DeiT-S & 45M & 3.6 & 32.1 & 47.1 & 51.7 & 56.6 \\ 
\qquad + SWRL  & & & 4.9(/20s)& \rednm{39.2} & \rednm{54.4} & \rednm{56.0} & 56.0 \\ \hdashline
\textbf{X$^2$Localizer (ours)} & DeiT-S & 45M & 3.6 &  \textbf{37.5} & \textbf{55.2} & \textbf{60.5} & \textbf{66.3} \\ 
\qquad + SWRL  & & & 4.9(/20s) & \redbf{44.9} & \redbf{62.2} & \redbf{64.5} & 64.5 \\ 
\bottomrule[1.5pt]
\end{tabular}
}
\vspace{1mm}
\caption{Frame-level localization performance with different prefix budgets. Retrieval is performed for all frames after coarse retrieval, reported in Recall@$k$ (\%) and inference latency (ms/frame). Following GAReT~\cite{GAReT}, Recall@1\% at this stage uses the full tile gallery and equals candidate-gallery Recall@All. \textbf{Bold} denotes the best results under the same inference strategy, and results in \rednm{red} indicate performance improvements brought by SWRL.
}\label{tab:f2f}
\end{table}

\subsection{Random-start Recovery}
To evaluate recovery ability after interruption or localization failure, we conduct a random-start recovery experiment. 
For each video, we randomly sample a temporal position and initialize localization using only a short prefix starting from that position. 
We evaluate both prefix-based coarse retrieval and subsequent incremental frame-to-tile retrieval. 
This protocol is more challenging than the standard prefix setting because the system cannot assume that the video starts from the beginning of a trajectory, and the available observation may contain limited or visually ambiguous context.
Table~\ref{tab:random_start} shows that X$^2$Localizer consistently outperforms GAReT across all random-start budgets. 
The improvement is largest under the single-frame restart setting, where temporal context is almost absent. 
These results suggest that the proposed asymmetric cross-grained alignment learns representations that are less dependent on complete sequence context. 
By combining budget-aware global supervision with token-aggregated local evidence, X$^2$Localizer can rapidly re-establish reliable coarse candidates and improve subsequent frame-level localization after a restart.


\begin{table*}[t]
\centering
\scriptsize
\setlength{\tabcolsep}{3pt}

\begin{minipage}[t]{0.32\textwidth}
\centering
\begin{tabular}{lc}
\toprule[1.2pt]
\textbf{Method} & \textbf{R@1/5/10/1\%} \\
\midrule
\rowcolor{gray!15}\multicolumn{2}{l}{\textit{Prefix Coarse Retrieval}} \\
GAReT & 20.0/46.9/58.9/74.4 \\
X$^2$Localizer & \textbf{24.5}/\textbf{55.0}/\textbf{66.9}/\textbf{82.3} \\
\midrule
\rowcolor{gray!15}\multicolumn{2}{l}{\textit{Incremental Fine-Grained Retrieval}} \\
GAReT & 35.6/51.4/56.2/61.2\\
X$^2$Localizer & \textbf{39.0}/\textbf{57.5}/\textbf{63.1}/\textbf{69.3} \\
\bottomrule[1.2pt]
\end{tabular}

\vspace{0.5mm}
\textbf{(a)} Random 1 frame ($\tau=1$).
\end{minipage}
\hfill
\begin{minipage}[t]{0.32\textwidth}
\centering
\begin{tabular}{lc}
\toprule[1.2pt]
\textbf{Method} & \textbf{R@1/5/10/1\%} \\
\midrule
\rowcolor{gray!15}\multicolumn{2}{l}{\textit{Prefix Coarse Retrieval}} \\
GAReT & 29.1/60.5/71.7/84.4 \\
X$^2$Localizer & \textbf{32.7}/\textbf{65.2}/\textbf{76.2}/\textbf{89.3} \\
\midrule
\rowcolor{gray!15}\multicolumn{2}{l}{\textit{Incremental Fine-Grained Retrieval}} \\
GAReT & 41.5/61.0/67.6/74.5 \\
X$^2$Localizer & \textbf{43.1}/\textbf{64.0}/\textbf{70.9}/\textbf{78.5} \\
\bottomrule[1.2pt]
\end{tabular}

\vspace{0.5mm}
\textbf{(b)} Random 5s clip ($\tau=2$).
\end{minipage}
\hfill
\begin{minipage}[t]{0.32\textwidth}
\centering
\begin{tabular}{lc}
\toprule[1.2pt]
\textbf{Method} & \textbf{R@1/5/10/1\%} \\
\midrule
\rowcolor{gray!15}\multicolumn{2}{l}{\textit{Prefix Coarse Retrieval}} \\
GAReT & 36.6/70.6/79.9/90.4 \\
X$^2$Localizer & \textbf{38.3}/\textbf{73.2}/\textbf{83.6}/\textbf{93.0} \\
\midrule
\rowcolor{gray!15}\multicolumn{2}{l}{\textit{Incremental Fine-Grained Retrieval}} \\
GAReT & 43.9/66.2/74.1/83.1 \\
X$^2$Localizer & \textbf{44.8}/\textbf{68.4}/\textbf{76.9}/\textbf{86.1} \\
\bottomrule[1.2pt]
\end{tabular}

\vspace{0.5mm}
\textbf{(c)} Random 20s clip ($\tau=4$).
\end{minipage}
\vspace{1mm}
\caption{Random-start recovery. The system is initialized from a random frame or clip, and we evaluate both coarse and subsequent incremental frame-level retrieval.} \label{tab:random_start}
\end{table*}

\subsection{Long-distance Progressive Localization}
\begin{figure}[!t]
    \centering
    \includegraphics[width=1.0\linewidth]{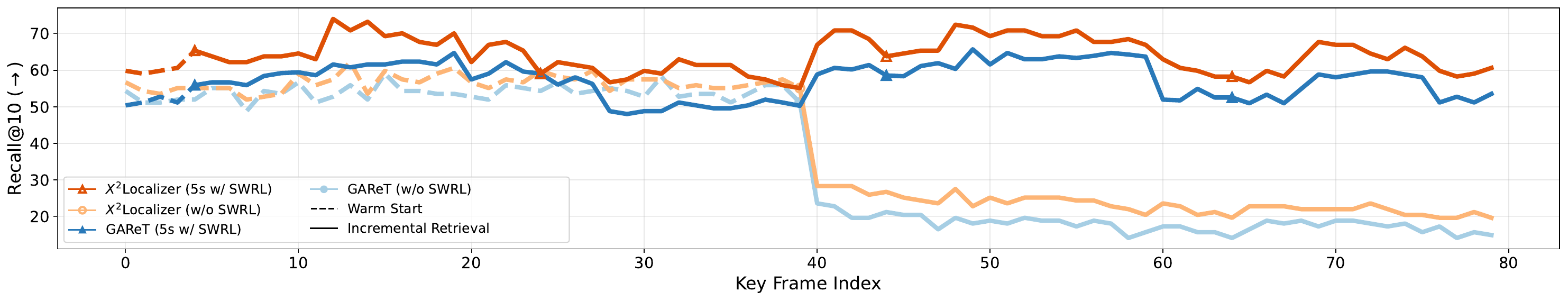}
    \vspace{-8mm}
    \caption{Long-range localization on the challenging subset. We compare our proposed method against GAReT, both with and without SWRL, which supports short warm-start and periodic re-localization via SWRL. Dashed lines indicate warm-up and backward refinement performance, whereas solid lines denote incremental frame-level refinement after activation.}
    \label{fig:longdistance}
\end{figure}

We further evaluate long-range deployment on the long-distance subset. 
In this setting, two temporally adjacent sequences from the same geographic source are concatenated to simulate continuous localization over an extended route. 
This setting is challenging because the initially retrieved aerial region may become outdated as the trajectory moves into a new area.
We compare four configurations: GAReT without SWRL, GAReT with SWRL, X$^2$Localizer without SWRL, and X$^2$Localizer with SWRL. 
Without SWRL, the system performs coarse retrieval once and then applies frame-level retrieval using the initial candidate gallery. 
With SWRL, the candidate region is periodically refreshed using a 5-second sliding-window prefix, enabling short warm-start and online re-localization. 
Frame-level retrieval is evaluated at key frames sampled approximately every second.
The performance curves in Fig.~\ref{fig:longdistance} reveal a limitation of the conventional full-video CVG paradigm. 
Although GAReT without SWRL performs well near the initial segment, its accuracy drops when the trajectory moves beyond the initially retrieved region. 
SWRL alleviates this issue by periodically updating the coarse candidate gallery, allowing the system to adapt to cross-region transitions without reprocessing the entire video. 
X$^2$Localizer further improves over GAReT under the same SWRL protocol, demonstrating that the proposed asymmetric cross-grained alignment provides more reliable short-prefix coarse retrieval. 
These results indicate that progressive re-localization and cross-grained alignment are complementary: SWRL supplies the online update mechanism, while X$^2$Localizer improves the quality of each short-budget re-localization step.

\begin{figure}[!t]
    \centering
    \includegraphics[width=1.0\linewidth]{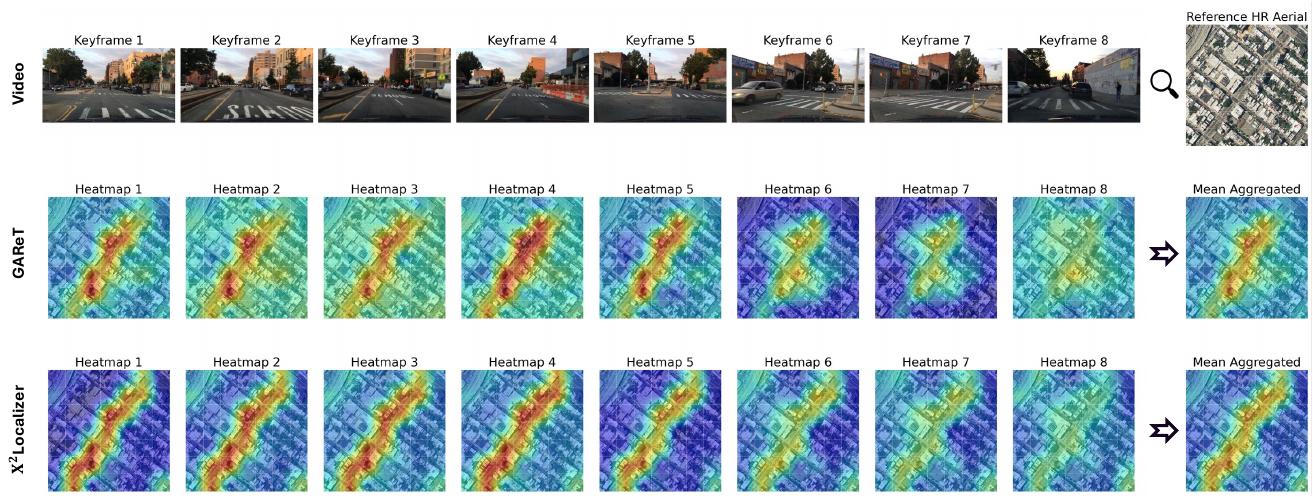}
    \vspace{-6mm}
    \caption{Qualitative visualization of frame-level localization cues. We show representative key frames, the corresponding aerial reference image, and the heatmaps produced by GAReT and X$^2$Localizer. X$^2$Localizer yields sharper and more temporally consistent responses across key frames, suggesting that asymmetric cross-grained alignment helps aggregate reliable local evidence for progressive localization.}
    \label{fig:visualexamples}
\end{figure}


\subsection{Ablation Study}

Table~\ref{tab:ablation_all}(a) validates the proposed asymmetric cross-grained objective. 
Using only $\mathcal{L}_{g}^{(\tau)}$ gives strong long-prefix performance but is weaker for short observations, while using only $\mathcal{L}_{f}^{(\tau)}$ improves the single-frame case but hurts longer budgets. 
This shows that global and fine-grained alignment are complementary. 
The symmetric variant with $\lambda_f=\lambda_g$ performs worse, especially at $\tau=8$, confirming the need for budget-dependent weighting. 
Removing either $\mathcal{L}_{\rm self}$ or $\mathcal{L}_{\rm teacher}$ reduces the average score, supporting both ranking regularizers.
Table~\ref{tab:ablation_all}(b) studies SWRL under the most challenging single-frame setting. 
Compared with fixed initial candidates, SWRL improves fine-grained retrieval by refreshing the candidate region with a sliding-window prefix. 
Using $s_{\mathrm{mix}}^{(\tau)}$ outperforms $s_g^{(\tau)}$, showing that cross-grained similarity produces better re-localization candidates. 
With $K_c=5$, SWRL improves R@1/5/10, while R@1\% slightly decreases.

\paragraph{Qualitative Analysis.}
Figure~\ref{fig:visualexamples} further visualizes the frame-level localization cues produced by GAReT and X$^2$Localizer. Compared with the baseline, X$^2$Localizer produces more concentrated and temporally consistent responses around the correct aerial regions. This is especially visible in visually ambiguous frames, where global video-level context alone may lead to diffuse or shifted activations. The visualization supports our quantitative findings: the proposed asymmetric cross-grained objective encourages the model to preserve local frame--tile evidence while maintaining stable prefix-level alignment, which is beneficial for progressive and short-budget localization.

\begin{table}[!t]
\centering
\scriptsize
\setlength{\tabcolsep}{3pt}

\begin{minipage}[t]{0.48\textwidth}
\centering
\begin{tabular}{l|cccc|c}
\toprule[1.2pt]
\textbf{Variant} & $\tau=1$ & $\tau=2$ & $\tau=4$ & $\tau=8$ & \textbf{Avg.} \\
\midrule
w/o X obj. & 16.9 & 25.9 & 41.4 & 50.2 & 33.6 \\
$\mathcal{L}_{\text{align}}=\mathcal{L}_{g}^{(\tau)}$ only & 20.3 & 28.6 & \textbf{43.2} & \underline{50.4} & \underline{35.6} \\
$\mathcal{L}_{\text{align}}=\mathcal{L}_{f}^{(\tau)}$ only & \underline{21.3} & 28.5 & 41.6 & \textbf{50.7} & 35.5 \\
$\lambda_{f}=\lambda_{g}$ (symmetric) & \underline{21.3} & 28.4 & 42.3 & 47.2 & 34.8 \\
w/o Distill $\mathcal{L}_{\text{self}}$ & 20.9 & \underline{28.7} & 42.4 & 49.8 & 35.5 \\
w/o Distill $\mathcal{L}_{\text{teacher}}$ & \textbf{21.6} & 28.6 & \underline{42.5} & 49.3 & 35.5 \\
\textbf{X$^2$Localizer} (X obj.) & \textbf{21.6} & \textbf{29.1} & \underline{42.5} & 50.3 & \textbf{35.9} \\
\bottomrule[1.2pt]
\end{tabular}

\vspace{0.5mm}
\textbf{(a)} Cross-grained objective (X obj.).
\end{minipage}
\hfill
\begin{minipage}[t]{0.48\textwidth}
\centering
\begin{tabular}{lcc|c|c}
\toprule[1.2pt]
\textbf{Variant} & \textbf{Sim.} & $K_c$ & \textbf{Gate@$K_c$} & \textbf{R@1/5/10/All} \\
\midrule
w/o SWRL      & $s_{\mathrm{mix}}^{(\tau)}$    & 5   & 50.9 & 34.3/48.7/52.5/\textbf{55.3} \\
SWRL          & $s_{\mathrm{mix}}^{(\tau)}$    & 5   & 50.9 & \textbf{41.4}/\textbf{53.5}/\textbf{53.5}/53.5 \\ \hdashline
w/o SWRL      & $s_{\mathrm{mix}}^{(\tau)}$    & 10  & 63.7 & 37.5/55.2/60.5/\textbf{66.3} \\
SWRL          & $s_g^{(\tau)}$ & 10  & 61.3 & 43.4/60.2/62.3/62.3 \\
\textbf{SWRL} & $s_{\mathrm{mix}}^{(\tau)}$    & 10  & 63.7 & \textbf{44.9}/\textbf{62.2}/\textbf{64.5}/64.5 \\
\bottomrule[1.2pt]
\end{tabular}

\vspace{0.5mm}
\textbf{(b)} Similarity choice and SWRL design when $\tau=1$.
\end{minipage}

\vspace{2mm}
\caption{
Ablation studies of X$^2$Localizer. 
(a) Component analysis of the asymmetric cross-grained alignment objective under different temporal budgets in Recall@1. 
(b) Inference-time analysis of the SWRL strategy under the single-frame setting ($\tau=1$). 
Gate@$K_c$ denotes whether the ground-truth global aerial region is included in the top-$K_c$ coarse candidates.
}
\label{tab:ablation_all}
\end{table}

\section{Conclusion}
In this work, we revisit cross-view video geo-localization and reformulate it as Progressive Cross-view Video Geo-localization (PCVG), a deployment-oriented setting that requires localization under varying temporal budgets, arbitrary starting positions, and long-range continuous streams. 
To address this task, we propose X$^2$Localizer, which learns robust prefix representations through an asymmetric cross-grained alignment objective. 
The objective jointly supervises global prefix-to-aerial retrieval and token-aggregated frame--aerial-tile matching, while adapting their relative weights according to the available temporal context. 
We further introduce ranking distillation to preserve full-video retrieval structure and SWRL to periodically refresh candidate regions during online inference. 
Experiments show that X$^2$Localizer remains competitive under the conventional full-video protocol while substantially improving short-prefix, random-start, and long-distance progressive localization. 
These results demonstrate the importance of combining budget-aware cross-grained alignment with online re-localization for practical cross-view video geo-localization.

\clearpage
\section*{Acknowledgment}
This work was mainly supported by the Engineering and Physical Sciences Research Council through an industrial CASE studentship with Ordnance Survey (Grant number EP/W522077/1 and EP/X524840/1). This work was supported in part by the National Natural Science Foundation of China under Grant No. 62503166, in part by the Hunan Provincial Research and Development Project under Grant number 2026QK3018, in part by the Yuelushan Industrial Innovation Center, in part by the Helmholtz Association of German Research Centers, in part by the Ministry of Science, Research and the Arts of Baden-W\"urttemberg (MWK) through the Cooperative Graduate School Accessibility through AI-based Assistive Technology (KATE) under Grant BW6-03, and in part by the Helmholtz Association Initiative and Networking Fund on the HAICORE@KIT and HOREKA@KIT partitions. 

\bibliography{egbib}

\renewcommand{\thesection}{\Alph{section}}
\renewcommand{\thesubsection}{\thesection.\arabic{subsection}}
\clearpage
\appendix

\section{Additional Methodology}
\label{app:method_details}

\subsection{Image-level Cross-view Pretraining}
\label{app:pretrain_loss}

Before adapting the model to video/global-aerial retrieval, we initialize the two image-level encoders with frame-tile (GPS-centered) supervision following previous works~\cite{TransGeoCG,GAReT}. Given a mini-batch of $B$ matched pairs $\{(\bm f_i,\bm a_i)\}_{i=1}^{B}$, we compute the ground-to-aerial similarity as
\begin{equation}
s_{\mathrm{img}}[i,j]
=
\operatorname{cos}
\left(
\phi_v(\bm f_i),
\phi_a(\bm a_j)
\right).
\end{equation}
The encoders are optimized with a ground-to-aerial soft-margin contrastive objective:
\begin{equation}
\mathcal{L}_{\mathrm{smcl}}
=
\frac{1}{B}
\sum_{i=1}^{B}
\log
\left(
1+
\sum_{j\neq i}
\exp
\left(
\frac{s_{\mathrm{img}}[i,j]-s_{\mathrm{img}}[i,i]}{\tau_{\mathrm{sm}}}
\right)
\right),
\end{equation}
where $\tau_{\mathrm{sm}}$ denotes the temperature. 
This stage learns an image-level cross-view representation by pulling each ground frame toward its GPS-aligned aerial tile while contrasting it against other aerial tiles in the mini-batch.

\subsection{GeoAdapter}
\label{app:geoadapter}

\begin{figure}[t]
    \centering
    \includegraphics[width=0.6\linewidth]{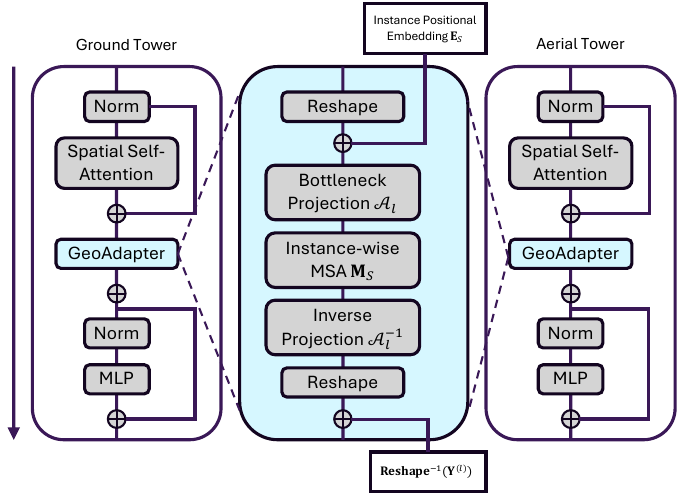}
    \caption{
    Architecture of GeoAdapter. }
    \label{fig:GeoAdpater}
\end{figure}

GeoAdapter~\cite{GAReT,STAdapter,AIM} shown in Fig.~\ref{fig:GeoAdpater} adapts the pretrained image encoders to sequence-level video/global-aerial matching without discarding the spatial representation learned during image-level pretraining. 
Instead of applying a temporal module after extracting frame embeddings, GeoAdapter is inserted inside the DeiT Transformer blocks. 
The pretrained spatial backbone is frozen, while the adapter branch and instance positional embeddings are trainable.

For each image instance, DeiT forms a token sequence
\begin{equation}
[
\mathbf{t}_{\mathrm{cls}},
\mathbf{t}_{\mathrm{dist}},
\mathbf{p}_1,\dots,\mathbf{p}_{N_p}
],
\end{equation}
where $\mathbf{t}_{\mathrm{cls}}$ and $\mathbf{t}_{\mathrm{dist}}$ are the classification and distillation tokens, and $\{\mathbf{p}_m\}_{m=1}^{N_p}$ are patch tokens. 
The ground branch treats video frames as instances, while the aerial branch treats spatial tiles of the global aerial image as instances. 
Let $S$ be the number of instances, with $S=\tau$ for a video prefix and $S=L$ for a global aerial image partitioned into $L$ tiles.

At the $l$-th block, spatial self-attention is first applied within each instance:
\begin{equation}
\mathbf{Z}^{(l)}
=
\mathbf{X}^{(l-1)}
+
\operatorname{MSA}_{\mathrm{sp}}^{(l)}
\left(
\operatorname{LN}
\left(
\mathbf{X}^{(l-1)}
\right)
\right).
\end{equation}
The intermediate tokens are then reshaped so that tokens at the same position can exchange information across the instance dimension:
\begin{equation}
\bar{\mathbf{Z}}^{(l)}
=
\operatorname{Reshape}
\left(
\mathbf{Z}^{(l)}
\right)
\in
\mathbb{R}^{B(N_p+2)\times S\times C}.
\end{equation}
GeoAdapter performs instance-wise attention through a lightweight bottleneck branch:
\begin{equation}
\mathbf{Y}^{(l)}
=
\mathcal{A}_{l}^{-1}
\left(
\operatorname{MSA}_{\mathrm{inst}}^{(l)}
\left(
\operatorname{LN}
\left(
\mathcal{A}_{l}
\left(
\bar{\mathbf{Z}}^{(l)}+\mathbf{E}_{S}
\right)
\right),
\mathbf{M}_{S}
\right)
\right),
\end{equation}
where $\mathcal{A}_{l}$ and $\mathcal{A}_{l}^{-1}$ are the adapter projections, $\mathbf{E}_{S}$ encodes the instance order or grid position, and $\mathbf{M}_{S}$ is an optional attention mask. 
For video inputs, $\mathbf{E}_{S}$ it is a learned flattened positional embedding.
The aerial mask disables cross-tile attention for patch tokens while allowing classification and distillation tokens to attend across tiles.

The adapter output is added back through a residual connection:
\begin{equation}
\tilde{\mathbf{X}}^{(l)}
=
\mathbf{Z}^{(l)}
+
\operatorname{Reshape}^{-1}
\left(
\mathbf{Y}^{(l)}
\right),
\end{equation}
followed by the standard Transformer MLP. 
After the final block, the projected classification and distillation tokens of each instance are averaged to form an instance embedding. 
The video branch outputs a global prefix embedding and frame-token embeddings, while the aerial branch outputs a global aerial embedding and tile-token embeddings:
\begin{equation}
\Phi_v(V^{(\tau)})
=
\left(
\mathbf{v}^{(g,\tau)}, 
\mathbf{P}^{(\tau)}
\right),
\qquad
\Phi_a(A^{\mathrm{global}})
=
\left(
\mathbf{a}^{(g)}, 
\mathbf{Q}
\right).
\end{equation}
The global embeddings support coarse prefix-to-aerial retrieval, and the token embeddings are used for fine-grained frame--tile alignment.

\section{More Implementation Details}
\label{app:implementation}

\paragraph{Architecture and Inputs}
We use DeiT-Small Patch16~\cite{DeiT} as the backbone for both towers. 
The ground-view and aerial towers share the same architecture but have separate parameters. 
Each backbone contains 12 Transformer blocks, 6 attention heads, and hidden dimension 384. 
The retrieval embedding dimension is set to 1000. 
Ground-view frames are resized to $216\times384$, and aerial tiles are resized to $256\times256$, with ImageNet normalization applied to both views.
Each video is represented by 8 uniformly sampled keyframes, and each global aerial image is divided into a $7\times7$ grid of non-overlapping tiles. 
Thus, the ground branch receives up to 8 frame instances, while the aerial branch receives 49 tile instances.
As the main comparison method, GAReT~\cite{GAReT} is trained on the same architecture but with different objectives.

\paragraph{Training}
The model is trained in two adaptation stages after image-level pretraining. 
In the first stage, the pretrained spatial backbone is frozen and the GeoAdapter modules are warmed up with full-video global retrieval only, \ie, 
$\mathcal{T}=\{8\}$ and $(\lambda_g^{(8)},\lambda_f^{(8)})=(1.0,0.0)$.
This stage runs for at most 50 epochs with early stopping patience 10, using full-video validation Recall@1 for model selection. The selected checkpoint initializes the progressive model and serves as the frozen teacher.

The progressive stage uses prefix budgets $\mathcal{T}=\{1,2,4,8\}$. 
The loss weights are summarized in Tab.~\ref{tab:app_training_weights}. 
Teacher distillation is applied to the global similarity matrix, and the aerial encoder is frozen during this stage. 
The model is trained for at most 50 epochs with early stopping patience 10.

\begin{table}[ht]
\centering
\begin{tabular}{c|c|c|c}
\toprule
Prefix $\tau$ & $\gamma_\tau$ & $\lambda_g^{(\tau)}$ & $\lambda_f^{(\tau)}$ \\
\midrule
1 & 0.05 & 1.0 & 2.0 \\
2 & 0.10 & 1.0 & 1.0 \\
4 & 0.25 & 1.0 & 0.5 \\
8 & 2.00 & 1.0 & 0.0 \\
\bottomrule
\end{tabular}
\caption{Weights used for progressive asymmetric training.}
\label{tab:app_training_weights}
\end{table}

Both stages use Adam~\cite{Adam} with learning rate $1\times10^{-4}$, weight decay 0, batch size 8, and 16-bit mixed precision. 
The contrastive and distillation temperatures are set to 0.07, and the token soft-aggregation temperature is set to 0.01. 
The prefix-to-full self-distillation and early-teacher distillation weights are $\eta_{\rm self}=0.2$ and $\eta_{\rm teacher}=1.0$, respectively. 
All experiments are conducted on a single NVIDIA RTX PRO 6000 GPU with 96GB memory.

\paragraph{Inference and Metrics}
For coarse retrieval, we evaluate prefix budgets $\tau=\{1,2,4,8\}$ and report Recall@1, Recall@5, Recall@10, and Recall@1\%. 
On the validation gallery, Recall@1\% corresponds to Recall@31. 
Unless otherwise specified, we evaluate both global similarity and mixed global--fine similarity.

For frame-level localization, the top-10 global aerial candidates are expanded into a frame-level aerial gallery and reranked with the image-level DeiT retriever. 
Follow GAReT~\cite{GAReT}, 1\% cutoff covers the entire candidate gallery and is therefore equivalent to candidate-gallery Recall@All.
A prediction is considered correct when its GPS distance to the ground-truth frame location is below 80 meters. 
For SWRL, candidate regions are refreshed every 20 seconds. 
In the 8-keyframe setting, this corresponds to evaluating the initial window and a later re-localization window beginning from the fifth sampled frame.

\section{Dataset and Evaluation Protocol}
\label{app:dataset_protocol}

\paragraph{Benchmark Split}
\label{app:dataset_split}

We evaluate on the standard cross-view video geo-localization benchmark introduced by GAMa~\cite{GAMa} and later used by GAReT~\cite{GAReT}. 
The training split contains 21,144 video-global aerial pairs and approximately 790K frame-level ground-aerial correspondences. 
The validation split contains 3,103 videos and approximately 116K frame-level correspondences. 
Since prior works GAMa~\cite{GAMa} and GAReT~\cite{GAReT} do not fully specify all preprocessing choices, we use one consistent preprocessing protocol across all experiments.

Following the GAReT-style video construction, each ground-view video is represented by 8 uniformly sampled keyframes. 
The same sampled representation is used for image-level pretraining, video/global-aerial adaptation, and coarse retrieval evaluation. 
The PCVG temporal budgets are therefore defined over this sampled sequence: $\mathcal{T}=\{1,2,4,8\}$, which correspond to the first frame, approximately 5 seconds, approximately 20 seconds, and the complete video observation.
In addition to the GPS-centered aerial tiles for pretraining, each global aerial image is partitioned into a $7\times7$ grid of non-overlapping and non-GPS centered tiles. 
The resulting 49 aerial tiles are used as the instance inputs of the aerial branch for adaptation. 
Frame-level supervision uses the GPS-aligned aerial tile associated with each sampled ground frame.

\paragraph{Progressive Retrieval}
\label{app:progressive_protocol}

The conventional CVG protocol evaluates retrieval after observing the complete video. 
PCVG instead evaluates the same gallery under multiple temporal budgets. 
Given a query prefix $V^{(\tau)}$, the model retrieves its corresponding global aerial image from the validation gallery. 
This protocol measures whether a model can produce reliable coarse localization before the full video is available.
For frame-level localization, the coarse retrieval result defines a candidate region set. 
The top-10 global aerial candidates are expanded into their associated frame-level aerial gallery, and frame-to-tile reranking is then performed inside this reduced search space. 
The frame-level gallery follows the same sampled frame-level correspondence protocol used by the benchmark. 
This evaluates whether early coarse retrieval provides a sufficiently accurate region for subsequent fine localization.

\paragraph{Random-start Recovery}
\label{app:random_start_protocol}

The random-start protocol evaluates re-localization from arbitrary temporal positions. 
For each validation clip, we sample a start index $r$ and form the query prefix
\begin{equation}
V_{r}^{(\tau)}
=
\{\bm f_{r}, \bm f_{r+1}, \dots, \bm f_{r+\tau-1}\}.
\end{equation}
The sampled start is constrained so that the evaluated prefix remains within the 8-frame sequence. 
All methods are evaluated with the same random seed (42) and start positions.
This protocol models delayed initialization, temporary tracking failure, or recovery after an interrupted observation. 
After coarse re-localization, frame-level retrieval is evaluated on subsequent observations from the sampled start position.

\paragraph{Long-range Progressive Localization}
\label{app:long_range_protocol}

\begin{figure}[t]
    \centering
    \includegraphics[width=\linewidth]{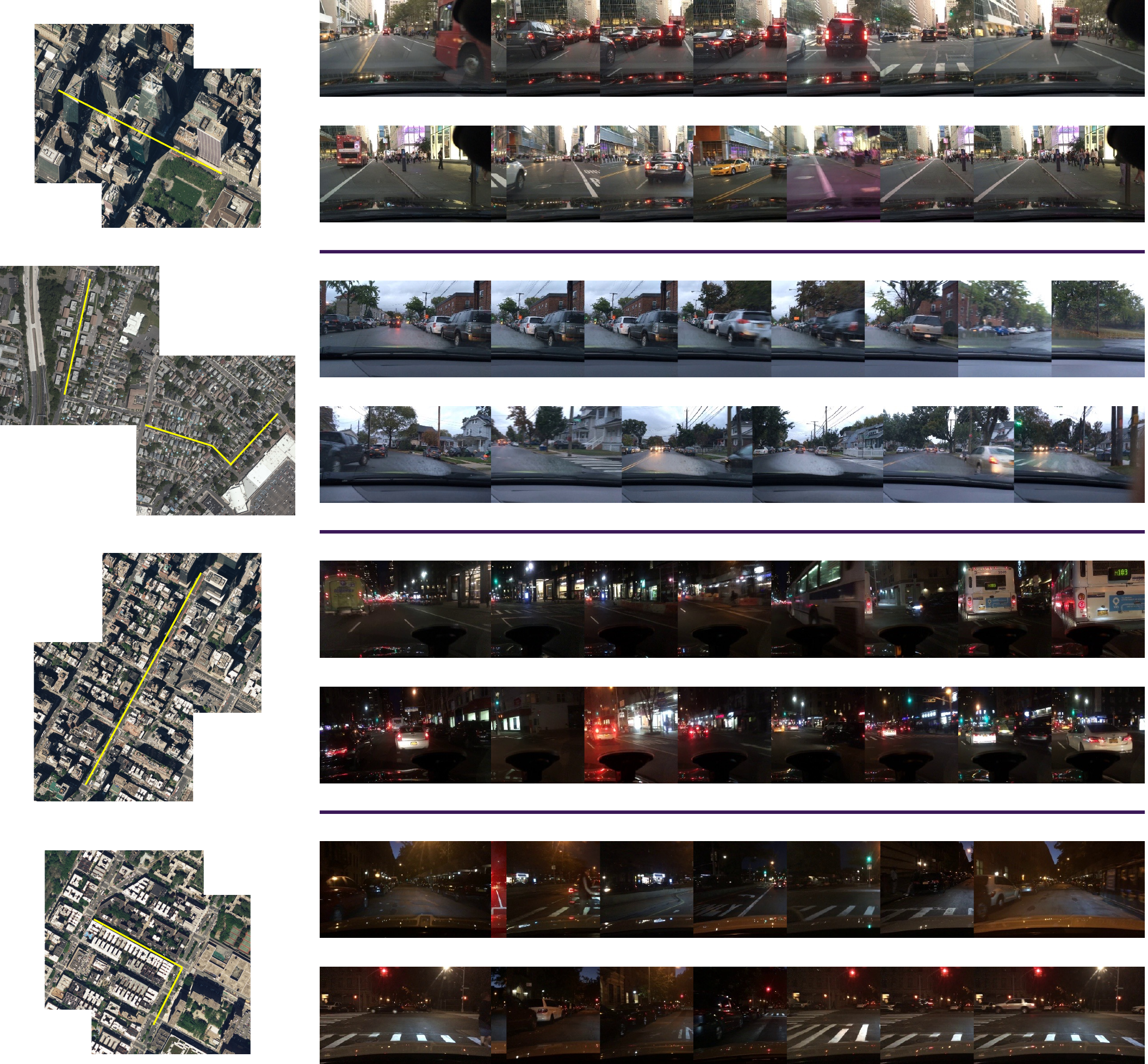}
    \caption{Examples from the constructed long-range validation subset. }
    \label{fig:longsubset}
\end{figure}

Standard validation clips evaluate localization within a short fixed temporal extent. 
To study longer deployments, we construct a long-distance subset by linking temporally adjacent validation clips from the same continuous driving trajectory. 
Only clips with valid frame-level correspondences are retained. 
Some linked clips contain temporal gaps of up to two minutes; we intentionally retain them, since such gaps reflect practical recovery scenarios where the localization system may be unavailable or unreliable for a period of time.
This construction yields 127 long sequences from 264 source clips, including 118 two-clip sequences, 8 three-clip sequences, and 1 four-clip sequence. 
For three-clip and four-clip sequences, we evaluate only the first two-clip segment in this paper. Figure~\ref{fig:longsubset} shows some examples, including daytime, nighttime, and break-refresh conditions. It is worth noting that although we used \textit{day split} from GAMa~\cite{GAMa}, a small amount of nighttime data was still included; therefore, we also took this nighttime data into account in the long-range subset.
On this subset, the query moves beyond a single isolated clip, and the initial candidate region may no longer remain valid throughout the sequence. 
SWRL addresses this setting by periodically refreshing the coarse candidate region using the most recent temporal window, followed by frame-level reranking within the refreshed gallery. 
The protocol therefore evaluates whether progressive localization remains stable over extended trajectories and after potential recovery events.

\section{More Experiments and Further Discussion}
\label{app:Aba}

\subsection{Long-range localization.}

\begin{figure}
    \centering
    \includegraphics[width=1.0\linewidth]{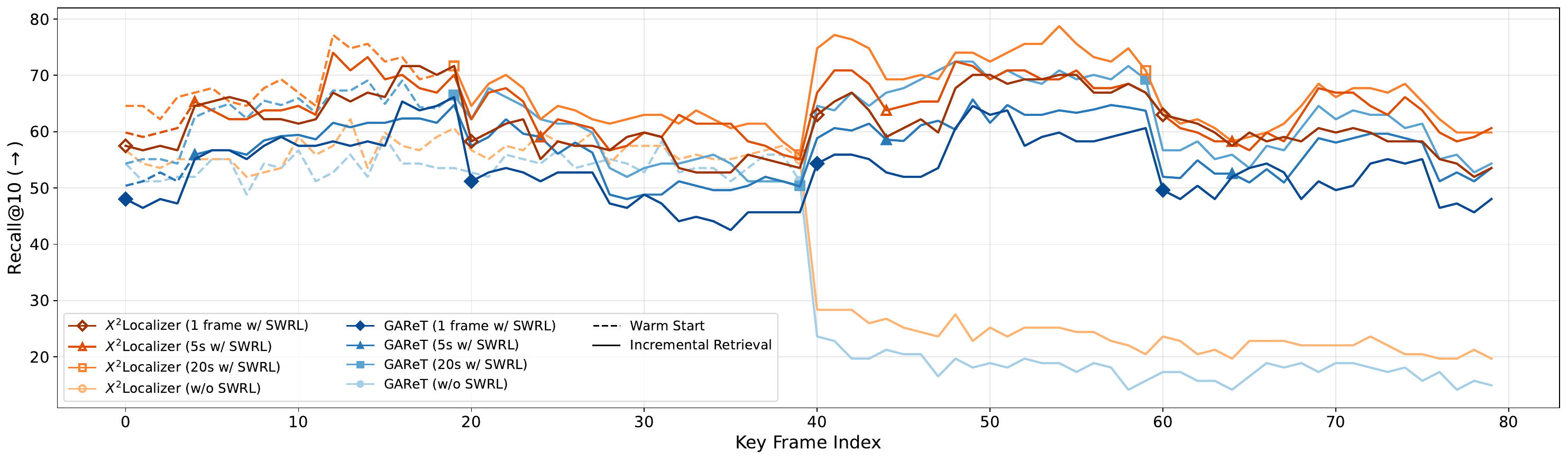}
    \caption{
    Long-range localization on the challenging subset. We compare X$^2$Localizer and GAReT under different SWRL re-localization budgets (1 frame, approximately 5 seconds, and approximately 20 seconds), together with the no-SWRL baseline. Vertical drops correspond to candidate-gallery refresh events. Longer windows generally improve re-localization reliability, while X$^2$Localizer consistently achieves stronger recovery under short observation budgets.
    }
    \label{fig:longdistance_app}
\end{figure}

Figure~\ref{fig:longdistance_app} further analyzes long-range deployment under different SWRL re-localization budgets. Here, we additionally evaluate 1-frame and 20-second re-localization windows. Several observations can be made. First, periodic candidate refresh is essential for extended trajectories. Without SWRL, both GAReT and X$^2$Localizer gradually degrade after the query moves beyond the initially retrieved region, since the candidate gallery remains fixed throughout the sequence. The performance drop becomes particularly evident after major trajectory transitions, where the original coarse retrieval is no longer valid. Second, longer re-localization windows generally provide more reliable recovery. For both methods, the 20-second SWRL configuration consistently achieves the highest Recall@10, indicating that additional temporal context improves coarse re-localization quality. However, this improvement comes at the cost of increased observation latency before a refresh can be triggered. Third, the proposed X$^2$Localizer exhibits a clear advantage in the low-latency regime. Under the extremely short 1-frame budget, X$^2$Localizer maintains competitive performance and substantially outperforms GAReT throughout most of the trajectory. The gap is especially visible immediately after refresh points, suggesting that asymmetric cross-grained alignment enables more discriminative coarse retrieval even from highly limited observations. Finally, the 5-second configuration offers a favorable trade-off between responsiveness and retrieval accuracy. It achieves performance close to the 20-second setting while reacting much faster to region transitions, making it a practical choice for online deployment. Overall, these results demonstrate that SWRL and X$^2$Localizer address complementary aspects of long-range localization. SWRL provides the mechanism for continuous candidate-region refresh, whereas X$^2$Localizer improves the quality of each refresh step, particularly when only short observation windows are available.

\subsection{Further Ablation Studies}
We provide additional ablations on the training weights, temperature parameters, inference similarity, and SWRL candidate refresh strategy. 
All results are reported on the validation split under the same PCVG protocol as in the main paper.

\begin{table}[t]
\centering
\scriptsize
\setlength{\tabcolsep}{4pt}
\begin{tabular}{llccccc}
\toprule
\textbf{Group} & \textbf{Setting} 
& $\tau=1$ & $\tau=2$ & $\tau=4$ & $\tau=8$ & \textbf{Avg.} \\
\midrule
\multirow{4}{*}{$\lambda_f$ for $\tau=\{1,2,4\}$}
& $(1,0.5,0.25)$           & 21.3 & 28.2 & 42.6 & 49.5 & 35.4 \\
& $(1,1,1)$                & 21.0 & 28.8 & \textbf{42.7} & \textbf{50.3} & 35.7 \\
& $(4,2,1)$                & 21.4 & \textbf{29.3} & 41.9 & 49.3 & 35.5 \\
& $\rednm{(2,1,0.5)}$      & \textbf{21.6} & 29.1 & 42.5 & \textbf{50.3} & \textbf{35.9} \\
\midrule
\multirow{3}{*}{$\tau_f$ for $\bm{s}_{\star}$}
& $0.005$                  & 21.4 & \textbf{29.2} & \textbf{42.5} & 50.2 & 35.8 \\
& $\rednm{0.01}$           & 21.6 & 29.1 & 42.5 & 50.3 & 35.9 \\
& $0.02$                   & 21.6 & \textbf{29.2} & 42.4 & 50.2 & \textbf{35.9} \\
\midrule
\multirow{3}{*}{$\tau_c$ for $\mathcal{L}_{\mathrm{CE}}$}
& $0.05$                   & 21.3 & 28.9 & \textbf{42.5} & \textbf{50.3} & 35.8 \\
& $\rednm{0.07}$           & 21.6 & 29.1 & \textbf{42.5} & \textbf{50.3} & \textbf{35.9} \\
& $0.10$                   & \textbf{21.9} & \textbf{29.2} & \textbf{42.5} & 50.1 & \textbf{35.9} \\
\bottomrule
\end{tabular}
\caption{
Sensitivity to asymmetric training weights and temperature parameters. 
We report Recall@1 (\%) for each temporal budget, and Avg. denotes the mean across budgets. 
Red indicates the default setting used by X$^2$Localizer.}
\label{tab:hyper_sensitivity}
\end{table}

\begin{table}[t]
\centering
\scriptsize
\setlength{\tabcolsep}{3pt}
\begin{tabular}{lcccccc}
\toprule
\textbf{Group} 
& \textbf{Setting} 
& $\tau=1$ & $\tau=2$ & $\tau=4$ & $\tau=8$ 
& \textbf{Avg.} \\
\midrule
\multirow{3}{*}{$\eta_\mathrm{self}$ for $\mathcal{L}_\mathrm{self}$}
& 0.1 & 20.9 & 28.3 & 42.3 & 50.1 & 35.4 \\
& \rednm{0.2} & \textbf{21.6} & \textbf{29.1} & \textbf{42.5} & \textbf{50.3} & \textbf{35.9} \\
& 0.4 & 21.4 & 28.9 & 42.1 & 49.9 & 35.6 \\
\midrule
\multirow{3}{*}{$\eta_\mathrm{teacher}$ for $\mathcal{L}_\mathrm{teacher}$}
& 0.5 & 20.3 & 28.1 & 41.9 & 49.8 & 35.0 \\
& \rednm{1.0} & \textbf{21.6} & \textbf{29.1} & \textbf{42.5} & \textbf{50.3} & \textbf{35.9} \\
& 1.5 & 20.9 & 28.8 & 42.0 & 50.2 & 35.5 \\
\midrule
\multirow{3}{*}{$\tau_d$ for $\mathcal{D}_{\mathrm{rank}}$}
& 0.05 & 21.2 & 28.8 & \textbf{42.5} & 50.2 & 35.7 \\
& \rednm{0.0}7 & 21.6 & \textbf{29.1} & \textbf{42.5} & \textbf{50.3} & \textbf{35.9} \\
& 0.10 & \textbf{21.8} & \textbf{29.1} & 42.4 & 50.1 & \textbf{35.9} \\
\bottomrule
\end{tabular}
\caption{
Sensitivity to ranking distillation hyperparameters. 
We vary the weights of prefix-to-full self-distillation and early-teacher distillation, as well as the ranking-distillation temperature. 
Results are Recall@1 (\%) under coarse retrieval. 
\rednm{Red} indicates the default setting.}
\label{tab:distill_sensitivity}
\end{table}

\begin{table}[t]
\centering
\scriptsize
\setlength{\tabcolsep}{3pt}
\begin{tabular}{lcccc}
\toprule 
\textbf{Method} & \textbf{Sim.} 
&  \textbf{R@1} & \textbf{R@5} & \textbf{R@10} 
\\
\midrule
\rowcolor{gray!15}\multicolumn{5}{l}{$\tau=8$} \\
\multirow{2}{*}{X$^2$Localizer} & $s_{\mathrm{global}}^{(\tau)}$ & 49.4 & 82.7 & 90.2 \\
& $s_{\mathrm{mix}}^{(\tau)}$    & \textbf{50.3} & \textbf{83.9} & \textbf{91.0} \\
\midrule
\rowcolor{gray!15}\multicolumn{5}{l}{$\tau=4$} \\
\multirow{2}{*}{X$^2$Localizer} & $s_{\mathrm{global}}^{(\tau)}$ & 41.1 & 76.6 & 85.3 \\
& $s_{\mathrm{mix}}^{(\tau)}$    & \textbf{42.5} & \textbf{78.2} & \textbf{86.7} \\
\midrule
\rowcolor{gray!15}\multicolumn{5}{l}{$\tau=2$} \\
\multirow{2}{*}{X$^2$Localizer} & $s_{\mathrm{global}}^{(\tau)}$ & 27.4 & 59.7 & 70.8 \\
& $s_{\mathrm{mix}}^{(\tau)}$    & \textbf{29.1} & \textbf{61.9} & \textbf{72.8} \\
\midrule
\rowcolor{gray!15}\multicolumn{5}{l}{$\tau=1$} \\
\multirow{2}{*}{X$^2$Localizer} & $s_{\mathrm{global}}^{(\tau)}$ & 20.2 & 49.6 & 61.3 \\
& $s_{\mathrm{mix}}^{(\tau)}$    & \textbf{21.6} & \textbf{50.9} & \textbf{63.7} \\
\bottomrule
\end{tabular}
\caption{
Effect of inference similarity for coarse retrieval. 
The mixed similarity $s_{\mathrm{mix}}^{(\tau)}$ combines global prefix-to-aerial similarity with token-aggregated fine-grained similarity, and consistently improves retrieval across temporal budgets.
}
\label{tab:similarities}
\end{table}

\begin{table}[t]
\centering
\scriptsize
\setlength{\tabcolsep}{3pt}
\begin{tabular}{lcccccc}
\toprule
\textbf{Variant} & \textbf{Sim.} & \textbf{K} & \textbf{Gate@K} & \textbf{R@1} & \textbf{R@5} & \textbf{R@10} \\
\midrule
\rowcolor{gray!15}\multicolumn{7}{l}{$\tau=4$} \\
w/o SWRL      & $s_{\mathrm{mix}}^{(\tau)}$    & 5   & 78.2 & \textbf{48.9} & \textbf{70.3} & \textbf{76.8} \\
SWRL          & $s_{\mathrm{mix}}^{(\tau)}$    & 5   & 78.2 & 48.7 & 67.3 & 71.7 \\ \hdashline
w/o SWRL      & $s_{\mathrm{mix}}^{(\tau)}$    & 10  & 86.7 & 46.3 & \textbf{70.3} & \textbf{78.8} \\
\textbf{SWRL} & $s_{\mathrm{mix}}^{(\tau)}$    & 10  & 86.7 & \textbf{47.4} & 70.0 & 77.3 \\
\midrule
\rowcolor{gray!15}\multicolumn{7}{l}{$\tau=2$} \\
w/o SWRL      & $s_{\mathrm{mix}}^{(\tau)}$    & 5   & 61.9 & 40.7 & 57.6 & 62.2 \\
SWRL          & $s_{\mathrm{mix}}^{(\tau)}$    & 5   & 61.9 & \textbf{46.4} & \textbf{61.1} & \textbf{62.6} \\ \hdashline
w/o SWRL      & $s_{\mathrm{mix}}^{(\tau)}$    & 10  & 72.8 & 41.7 & 61.5 & 68.0 \\
\textbf{SWRL} & $s_{\mathrm{mix}}^{(\tau)}$    & 10  & 72.8 & \textbf{48.0} & \textbf{67.3} & \textbf{71.6}\\
\midrule
\rowcolor{gray!15}\multicolumn{7}{l}{$\tau=1$} \\
w/o SWRL      & $s_{\mathrm{mix}}^{(\tau)}$    & 5   & 50.9 & 34.3 & 48.7 & 52.5 \\
SWRL          & $s_{\mathrm{mix}}^{(\tau)}$    & 5   & 50.9 & \textbf{41.4} & \textbf{53.5} & \textbf{53.5} \\ \hdashline
w/o SWRL      & $s_{\mathrm{mix}}^{(\tau)}$    & 10  & 63.7 & 37.5 & 55.2 & 60.5 \\
\textbf{SWRL} & $s_{\mathrm{mix}}^{(\tau)}$    & 10  & 63.7 & \textbf{44.9} & \textbf{62.2} & \textbf{64.5} \\
\bottomrule[1.2pt]
\end{tabular}
\caption{
Effect of SWRL candidate refresh in the recovery setting. 
$K$ denotes the number of coarse candidate places used for frame-level reranking, and Gate@K measures whether the correct place is included in the candidate set. 
Frame-level recall is computed with the 80-meter GPS threshold.
}\label{tab:SWRL_aba}
\end{table}

\paragraph{Training hyperparameters.}
Table~\ref{tab:hyper_sensitivity} studies the sensitivity of X$^2$Localizer to the asymmetric fine-grained weights and temperature parameters. 
The model is stable across a broad range of settings: all tested configurations remain within a narrow performance band, and the default configuration achieves the best average Recall@1 across temporal budgets. 
The schedule $\lambda_f=(2,1,0.5)$ gives stronger supervision to shorter prefixes while gradually reducing fine-grained supervision as more temporal context becomes available, which provides the best balance across early and full-video retrieval. 
The token aggregation temperature $\tau_f$ and contrastive temperature $\tau_c$ show low sensitivity around the default values, indicating that the proposed objective is not heavily dependent on a single narrow hyperparameter choice.

\paragraph{Ranking distillation.}
Table~\ref{tab:distill_sensitivity} evaluates the hyperparameters of prefix-to-full self-distillation and early-teacher ranking distillation. 
A moderate self-distillation weight performs best: too weak a weight provides limited ranking regularization for short prefixes, while too strong a weight can over-constrain the prefix representation. 
Similarly, the teacher distillation weight $\eta_{\mathrm{teacher}}=1.0$ gives the best overall performance, confirming that the early full-video model provides a useful but not overly dominant ranking prior. 
The ranking-distillation temperature is also robust, with $\tau_d=0.07$ slightly preferred on average.

\paragraph{Inference similarity.}
Table~\ref{tab:similarities} compares the global similarity $s_{\mathrm{global}}^{(\tau)}$ and the mixed similarity $s_{\mathrm{mix}}^{(\tau)}$ at inference time. 
The mixed similarity consistently improves retrieval across all temporal budgets. 
The gains are especially useful under partial observations, where token-level frame--tile evidence complements the global prefix representation. 
For example, under $\tau=2$, using $s_{\mathrm{mix}}^{(\tau)}$ improves Recall@1 from 27.4 to 29.1 and Recall@10 from 70.8 to 72.8. 
This supports using global and fine-grained evidence jointly during progressive retrieval.

\paragraph{SWRL candidate refresh.}
Table~\ref{tab:SWRL_aba} evaluates the effect of SWRL in the recovery setting. 
SWRL is most beneficial for short prefixes and strict retrieval ranks, where the current observation is limited and the initial candidate region can be unreliable. 
With $K=10$, SWRL improves Recall@1 by 7.4 points for $\tau=1$ and 6.3 points for $\tau=2$. 
For longer prefixes, the static candidate set is already more reliable, and SWRL becomes comparable rather than uniformly better. 
This behavior is consistent with the role of SWRL: it is designed primarily for recovery and long-range deployment, where refreshing the candidate region is more important than maximizing broad candidate coverage at very large ranks.

\begin{figure}[t]
    \centering
    \includegraphics[width=\linewidth]{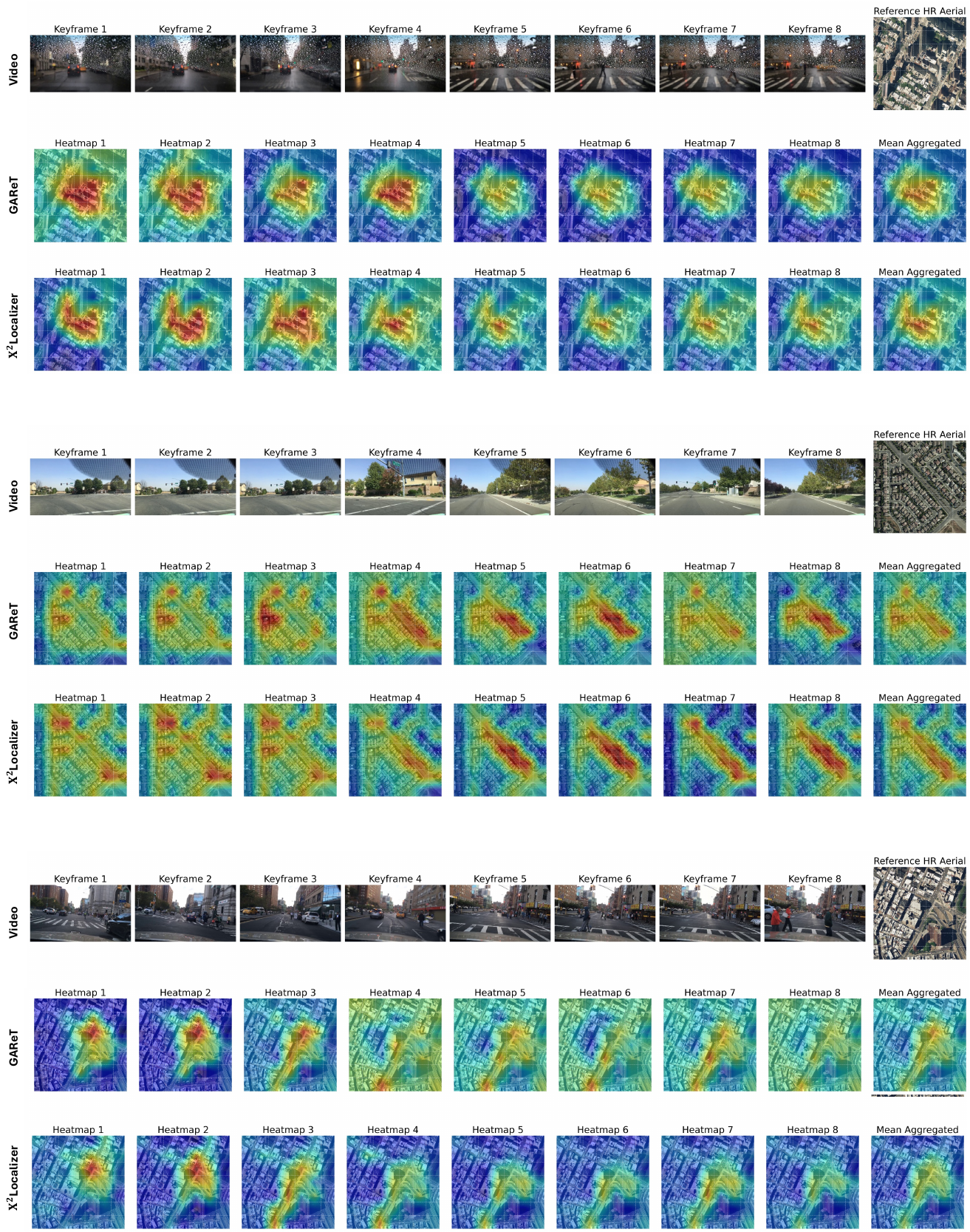}
    \caption{
    Additional qualitative heatmap comparisons between GAReT and X$^2$Localizer. 
Each example includes eight sampled video keyframes, the reference high-resolution aerial image, frame-level heatmaps, and the mean aggregated heatmap. 
X$^2$Localizer produces more coherent responses across keyframes, indicating improved temporal consistency of local evidence.
    }
    \label{fig:longvq1}
\end{figure}

\begin{figure}[t]
    \centering
    \includegraphics[width=\linewidth]{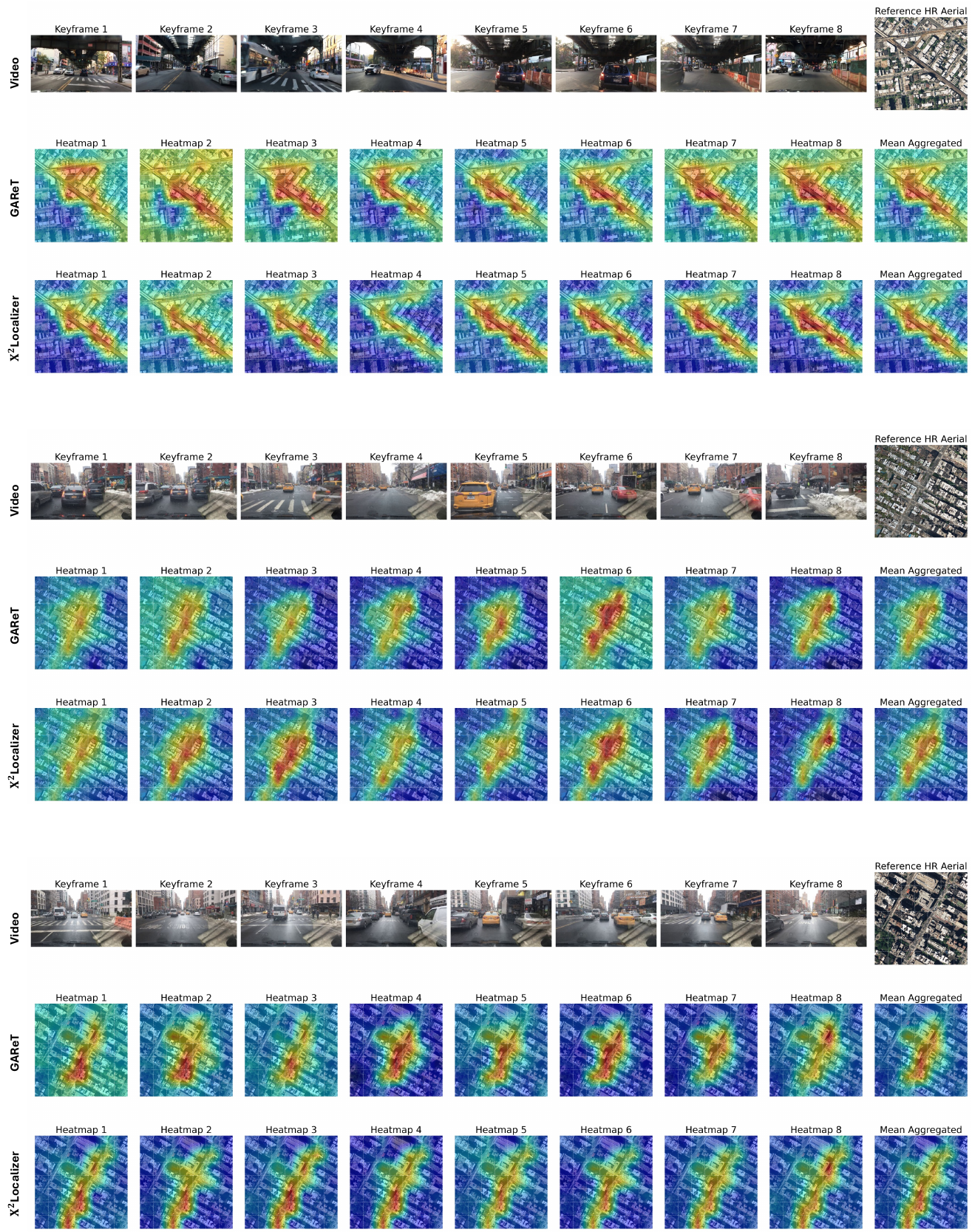}
    \caption{
More qualitative examples of frame-level localization heatmaps. 
Compared with GAReT, X$^2$Localizer tends to generate sharper and more spatially stable activations around road-like regions in the aerial image, supporting the effectiveness of token-aggregated frame--aerial-tile alignment.
    }
    \label{fig:longvq2}
\end{figure}

\paragraph{More Visual Examples.}
Figures~\ref{fig:longvq1} and~\ref{fig:longvq2}  provide additional qualitative comparisons between GAReT and X$^2$Localizer. 
For each example, we show the sampled video keyframes, the corresponding reference aerial image, per-frame heatmaps, and the mean aggregated heatmap. 
Across diverse scenes, X$^2$Localizer generally produces more concentrated and temporally consistent responses along plausible road structures, while GAReT often shows more diffuse or unstable activations. 
These examples complement the quantitative results by illustrating how asymmetric cross-grained alignment improves local evidence aggregation under progressive observation budgets.

\section{Limitations and Future Work}
Although X$^2$Localizer improves progressive and long-range cross-view video geo-localization, several limitations remain. 
First, the current benchmark follows the sampled-keyframe protocol, where each video is represented by 8 uniformly sampled frames. 
Future work may evaluate denser streaming inputs and adaptive frame selection. 
Second, SWRL refreshes candidate regions at a fixed interval and currently performs re-localization without explicitly reusing previous retrieval results. 
Future work may develop confidence-aware and history-aware refresh policies that reduce unnecessary global retrieval while improving recovery. 
Third, our framework mainly focuses on progressive coarse re-localization and does not explicitly model temporal consistency during frame-level decoding. 
This is complementary to the TransRetriever module in GAReT~\cite{GAReT}, which autoregressively decodes frame-level matches from top-$K$ candidates conditioned on previous predictions to maintain trajectory consistency. 
Integrating such temporal-consistency decoding with SWRL is a promising direction, where SWRL provides long-range candidate refresh while an autoregressive decoder stabilizes local frame-to-frame localization. 
Fourth, the current framework relies on visual evidence only. 
Incorporating inertial cues, map priors, or uncertainty estimation may further improve robustness in visually ambiguous regions, nighttime scenes, and long temporal gaps. 
Finally, we have not yet evaluated deployment on edge platforms. 
Future work may investigate model compression, feature caching, and hardware-aware inference for real-time onboard localization.

\end{document}